\documentclass{article} 
\usepackage{iclr2027_conference,times}

\usepackage{amsmath,amsfonts,bm}

\def\eqref#1{equation~\ref{#1}}

\def\1{\bm{1}}

\DeclareMathAlphabet{\mathsfit}{\encodingdefault}{\sfdefault}{m}{sl}
\SetMathAlphabet{\mathsfit}{bold}{\encodingdefault}{\sfdefault}{bx}{n}

\def\sC{{\mathbb{C}}}

\usepackage{hyperref}
\usepackage{url}
\usepackage[utf8]{inputenc} 
\usepackage[T1]{fontenc}    
\usepackage{enumitem}
\usepackage{booktabs}       
\usepackage{tabularx}
\usepackage{amsfonts}       
\usepackage{nicefrac}       
\usepackage{microtype}      
\usepackage{wrapfig}
\usepackage{makecell}
\usepackage{array}

\usepackage{graphicx}
\usepackage{amsthm}
\usepackage{xcolor,tikz,makecell,multirow}
\usepackage{colortbl}
\usepackage[most]{tcolorbox}

\definecolor{TakeawayHeader}{RGB}{50,50,50}
\definecolor{TakeawayBg}{RGB}{240,243,250}
\definecolor{MicrosoftBlue}{HTML}{0078D4}
\definecolor{TitleCardBg}{RGB}{246,248,251}
\definecolor{TitleCardRule}{RGB}{218,226,235}
\definecolor{NSTableStripe}{RGB}{231,240,249}
\definecolor{ResultHeader}{RGB}{236,236,236}
\definecolor{ResultSection}{RGB}{225,225,225}
\definecolor{ResultAlt}{RGB}{248,248,248}
\definecolor{ResultRule}{RGB}{170,170,170}
\newcommand{\nstablerules}{\arrayrulecolor{ActionBlue}\setlength{\arrayrulewidth}{0.45pt}}
\newcommand{\resulttablerules}{\arrayrulecolor{ResultRule}\setlength{\arrayrulewidth}{0.45pt}}
\newcommand{\normaltablerules}{\arrayrulecolor{black}\setlength{\arrayrulewidth}{0.4pt}}
\tcbset{
  takeawaybox/.style={
    enhanced,
    colback=TakeawayBg,
    colframe=TakeawayHeader,
    fonttitle=\bfseries\sffamily\small,
    coltitle=white,
    attach boxed title to top left={yshift=-2mm, xshift=4mm},
    boxed title style={colback=TakeawayHeader, rounded corners},
    rounded corners,
    boxrule=0.4pt,
    top=4mm, left=4pt, right=4pt, bottom=4pt,
  },
  promptbox/.style={
    enhanced,
    colback=blue!4!white,
    colframe=blue!20!white,
    rounded corners,
    boxrule=0.5pt,
    fontupper=\small\ttfamily,
    top=6pt, bottom=6pt, left=6pt, right=6pt,
  }
}

\definecolor{ActionBlue}{RGB}{0,112,192}
\definecolor{ActionYellow}{RGB}{255,192,0}
\definecolor{ActionGreen}{RGB}{0,176,80}
\definecolor{ActionOrange}{RGB}{233,113,50}
\definecolor{ActionRed}{RGB}{192,0,0}

\newcommand{\actioncircle}[3][2.75ex]{%
  \tikz[baseline=-0.6ex]\node[
    circle,
    fill=#2,
    minimum size=#1,
    inner sep=0pt,
    text=white,
    font=\sffamily
  ] {#3};}

\theoremstyle{definition}
\newtheorem{definition}{Definition}[section]

\title{Reinforcing Agentic Creativity in Scientific Ideation with Night Science}

\author{%
  Priyanka Kargupta\textsuperscript{1}\thanks{Work completed while interning at Microsoft. Corresponding authors: \texttt{pk36@illinois.edu}, \texttt{sjauhar@microsoft.com}}
  \And
  Silviu Cucerzan\textsuperscript{2}
  \And
  Shweti Mahajan\textsuperscript{3}
  \And
  Allen Herring\textsuperscript{2}
  \AND
  Jiawei Han\textsuperscript{1} \quad Ryen W. White\textsuperscript{2} \quad Sujay Kumar Jauhar\textsuperscript{2} \\ \\
  \noindent\parbox{\linewidth}{\centering\small \textsuperscript{1}University of Illinois Urbana-Champaign \quad \textsuperscript{2}Microsoft \quad \textsuperscript{3}Microsoft Research}
}

\newcommand{\papertitlecard}{%
\begin{center}
\begin{tcolorbox}[
  enhanced,
  width=\textwidth,
  colback=TitleCardBg,
  colframe=TitleCardRule,
  boxrule=0.4pt,
  arc=10pt,
  outer arc=10pt,
  left=14pt,
  right=14pt,
  top=12pt,
  bottom=12pt
]
\setlength{\parindent}{0pt}
\begin{minipage}[t]{\linewidth}
{\LARGE\sc Reinforcing Agentic Creativity in\\Scientific Ideation with Night Science\par}
\end{minipage}
\vspace{10pt}

{\normalsize
Priyanka Kargupta\textsuperscript{1,*} \quad
Silviu Cucerzan\textsuperscript{2} \quad
Shweti Mahajan\textsuperscript{3} \quad
Allen Herring\textsuperscript{2}\\[-1pt]
Jiawei Han\textsuperscript{1} \quad
Ryen W. White\textsuperscript{2} \quad
Sujay Kumar Jauhar\textsuperscript{2,*}\par}
\vspace{6pt}
{\footnotesize
\textsuperscript{1}University of Illinois Urbana-Champaign \quad
\textsuperscript{2}Microsoft \quad
\textsuperscript{3}Microsoft Research\par}
\vspace{4pt}
\noindent
\begin{minipage}[c]{0.77\linewidth}
{\footnotesize
\textbf{Correspondence:} \textcolor{MicrosoftBlue}{\texttt{pk36@illinois.edu}}, \textcolor{MicrosoftBlue}{\texttt{sjauhar@microsoft.com}}\par
\textbf{Blog:} \textcolor{MicrosoftBlue}{\href{http://pkargupta.github.io/night_scientist.html}{\texttt{pkargupta.github.io/night\_scientist}}}\\
\textbf{Code:} \textcolor{MicrosoftBlue}{\href{https://github.com/microsoft/ai_night_scientist}{\texttt{microsoft/ai\_night\_scientist}}}\\
\textsuperscript{*}\textit{Work completed while interning at Microsoft.}}
\end{minipage}
\hfill
\begin{minipage}[c]{0.18\linewidth}
\raggedleft
\includegraphics[height=0.27in]{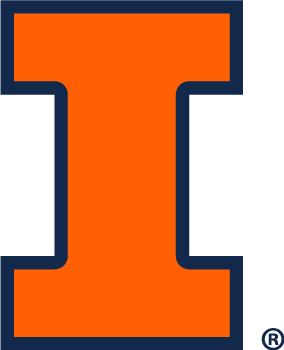}\hspace{9pt}
\includegraphics[height=0.27in]{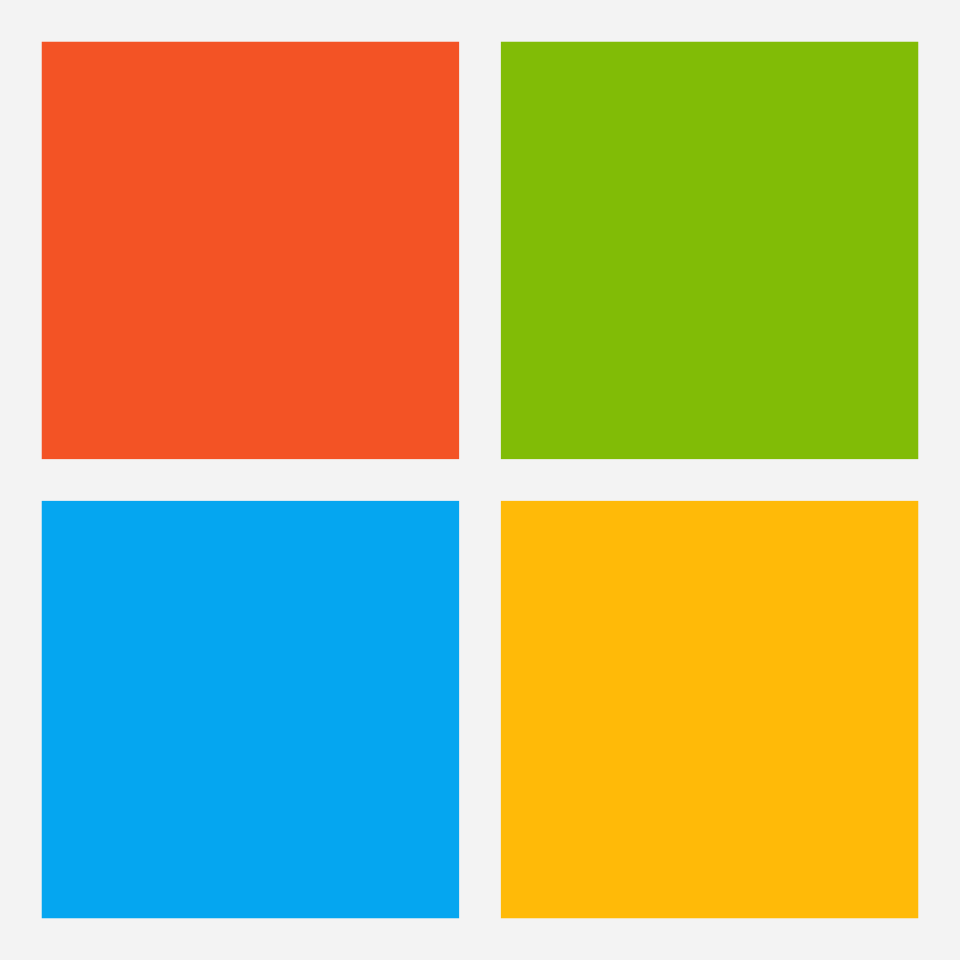}
\end{minipage}
\end{tcolorbox}
\end{center}
\vspace{-2pt}
}

\iclrfinalcopy 
\begin{document}

\papertitlecard

\begin{abstract}
Large language models (LLMs) excel at structured, verifiable tasks, but their low-entropy bias can produce homogeneous and predictable outputs, limiting their utility for open-ended scientific ideation. Effective discovery, however, spans a broader creative spectrum: from structured \textit{day science} to loosely structured, serendipitous \textit{night science} that reaches ideas beyond those typically considered. We introduce \textsc{\textbf{AI Night-Scientist}}, an agentic framework that uses reinforcement learning to teach models \emph{when} and \emph{how} to depart from predictable reasoning. Grounded in cognitive science, we model creativity along three axes: \textit{action} (what to do and how creatively), \textit{process} (when to explore versus exploit), and \textit{outcome} (the novelty and usefulness of the resulting idea). We use these axes to train models with GRPO, exposing them to varying degrees and forms of creativity throughout training. This produces substantially more diverse scientific proposals, expanding the range of research directions by 27.8\% and contribution types by 14.9\% over the base model. It also improves predicted citation impact by up to 32.0 percentage points and originality by 66.2 points. These gains cannot be reproduced by simply increasing decoding temperature; instead, we find that semantic guidance specifying \emph{what kind} of creativity to pursue is critical. Overall, our results suggest that \textbf{creativity is a learnable, multi-level ability that can be shaped to help researchers reach ideas beyond those typically explored by LLMs}.
\end{abstract}

\section{Introduction}

\par  Large language models (LLMs) have excelled at structured, systematic tasks with clear verifiability (e.g., coding and quantitative reasoning), where their performance is often improved with reinforcement learning (RL) \citep{guo2025deepseek, wang2025reinforcement, zhong2024can, liu2024exploring}. This success is consistent with a broader tendency toward minimizing token entropy \citep{agarwal2025unreasonable}, where models favor high-probability outputs that reflect frequent patterns and expected answers in training data \citep{mccoy2024embers}. While LLMs have increasingly been applied to scientific ideation \citep{chenglillmideas, gottweis2025towards}, they lack originality \citep{zhao2025assessing}, tend to generate homogeneous outputs \citep{wenger2025we},  and even plagiarize at nontrivial rates \citep{gupta2025all}. Ultimately, \textbf{this directly conflicts with the key attributes of open-ended, creative scientific discovery}: novelty, diversity, and serendipity.

\par Prior methods depict discovery as a highly structured process, involving hypotheses derived from prior observations and/or data, tested against evidence, and refined or rejected accordingly \citep{gottweis2025towards, agarwal2026autodiscovery}. But this is only *part* of the scientific discovery process. They typically view creativity as only an attribute of ideas rather than as part of the process, treating individual actions as fixed behaviors (e.g., search, write) and executing them through scaffolded pipelines \citep{gu2024llms, lu2024llm}. While this supports the critical reasoning crucial for validating existing ideas, this overlooks the creative reasoning necessary for discovering new ones. Both are complementary and crucial perspectives of scientific discovery, referred to as \textbf{\textit{day}} and \textit{\textbf{night science}} \citep{wechsler2018creative, halpern2007nature, stent1988statue, yanai2019night}.

\par Night science captures the often-neglected nature of human-driven discovery: loosely structured and dynamic exploration driven by highly creative actions, such as exploring distant analogies (e.g., biology-inspired technology), considering alternative perspectives after serendipitous encounters (e.g., debates with colleagues from other domains), and acting on partly-formalized intuitions (e.g., abandoning status quo assumptions). Together, day and night science form a spectrum that human researchers have smoothly traversed to uncover breakthroughs \citep{de2022reasoning, dwyer2025evaluation}, such as chemotherapy and penicillin \citep{hirsch2006anniversary,ligon2004penicillin}.

\begin{figure*}
    \centering
    \includegraphics[width=\linewidth]{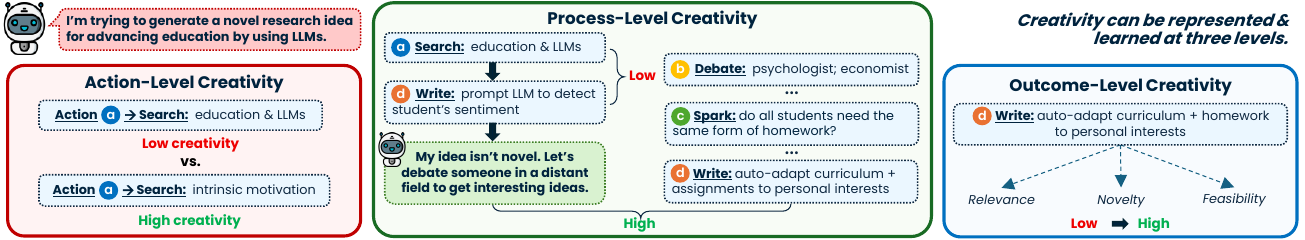}
    \caption{Given an input task, creativity can be injected into three different axes of reasoning: \textit{action, process}, and \textit{outcome}. Moreover, each level can be executed with \textit{varying degrees of creativity}.}
    \label{fig:three_levels}
    \vspace{-4mm}
\end{figure*}

\par We hypothesize that LLMs can better traverse this spectrum if given explicit control over \emph{when} and \emph{how} to deviate from high-probability output, enabling targeted doses of night science while maintaining day science's goal-directed behavior. To test this, we introduce \textsc{\textbf{AI Night-Scientist}}, an RL-based agentic framework that incentivizes models to exercise this control effectively. Grounded in cognitive science \citep{lubart2001models, cohen1989continuum, dwyer2025evaluation, harvey2023toward}, it embeds creativity along three axes of reasoning (Figure~\ref{fig:three_levels}):

\begin{itemize}[leftmargin=1em, itemsep=0.25em]
\item \textbf{\textit{Action-level} (\emph{how}):} Associates each action with a creativity level, allowing the same action to vary from conventional, high-probability behavior to more exploratory and unconventional behavior.

\item \textbf{\textit{Process-level} (\emph{when}):} Controls when to shift between lower- and higher-creativity actions based on how the multi-step trajectory unfolds.

\item \textbf{\textit{Outcome-level} (\emph{what}):} Captures the creativity of the resulting idea, favoring outputs that are novel while remaining relevant and useful.
\end{itemize}

\par We utilize this framework to train an LLM-based agent using GRPO \citep{shao2024deepseekmath} for generating \textit{scientific research proposals}, where identifying promising ideas often requires long-horizon creative reasoning beyond immediately verifiable evidence. Overall, our work argues that LLM-based support for scientific discovery should span the full day-to-night science spectrum. \textsc{AI Night-Scientist} shows that conventional LLMs do not naturally navigate this spectrum effectively, but targeted reinforcement learning can reshape when and how they depart from structured, predictable reasoning, leading to more creative outcomes. Our contributions can be summarized as:
\begin{enumerate}[leftmargin=1em, itemsep=0.25em]
    \item We introduce \textsc{\textbf{AI Night-Scientist}}, an RL-based agentic framework that explicitly represents creativity across actions, reasoning processes, and outcomes for long-horizon scientific ideation.
    \item We show that creativity is more than sampling stochasticity: explicit guidance on \emph{how} to be creative via RL helps an agent learn \emph{when} creative deviations are useful, while higher temperature does not.
    \item Empirically, \textsc{AI Night-Scientist} produces more diverse scientific proposals than its base model, expanding the range of research directions by 27.8\% and contribution types by 14.9\%, while improving predicted impact by up to 32.0 percentage points and originality by 66.2 points.
\end{enumerate}

\section{Background and Related Work}

\par Scientific discovery has been characterized as an interplay between structured, hypothesis-driven \textit{day science} and more exploratory, intuition- and serendipity-driven \textit{night science} \citep{stent1988statue, yanai2019night}. Classic theories of creativity characterize creative thought through remote associations between otherwise distant concepts~\citep{mednick1962associative}, generative and exploratory modes of cognition~\citep{ward1999creative}, and outcomes that are both original and useful~\citep{runco2012standard}. Creativity can also vary in degree: prior work describes a continuum of creative behavior~\citep{cohen1989continuum}, with problem-solving strategies ranging from paradigm-preserving to paradigm-stretching and paradigm-breaking~\citep{mcfadzean1998creativity}. Related accounts distinguish between exploring existing conceptual spaces and transforming them to enable qualitatively new ideas~\citep{boden1998creativity}, while theories of creative ideation show that originality can arise either by flexibly exploring many conceptual directions or by persistently exploring a few in greater depth~\citep{nijstad2010dual}. Together, these perspectives motivate our view of creativity as varying both \emph{where} it enters reasoning (within actions, processes, and outcomes) and \emph{how strongly} it is expressed.

\par This view contrasts with most current LLM-based approaches to scientific discovery. Existing research agents broaden ideation through retrieval, search, multi-agent interaction, and iterative generation~\citep{gu2024llms, lu2024llm, kargupta2025tree, gottweis2025towards}, but generally treat actions such as searching, debating, and writing as fixed behaviors and primarily assess creativity in the resulting idea. This leaves little control over \emph{how creatively} individual actions are performed or \emph{when} creative deviations should occur throughout reasoning. Relatedly, \citet{kargupta2025cognitive} find that LLMs struggle with the metacognitive awareness needed to monitor and adapt their reasoning, further limiting their ability to flexibly shift between critical and creative modes.

\par Recent work has begun targeting the mechanisms that produce creative scientific ideas. \citet{o2025sparks} use structured assumption inversion to generate novel hypotheses, inspiring the \textit{spark} action in our framework, while \citet{kargupta2025beyond} and \citet{kargupta2026sparking} use retrieval to identify research gaps and surface interdisciplinary inspiration. Other work instead learns scientific capabilities directly: GIANTS~\citep{he2026giants} trains smaller models to anticipate scientific insights, while \citet{tong2026ai} study whether models can learn scientific taste. Together, these approaches suggest that not only scientific outputs, but also the processes that produce them, can be shaped through structure and supervision. Reinforcement learning offers a way to shape this process without prescribing exactly how discovery should unfold, although open-ended ideation has no single correct answer and must balance qualities such as novelty, relevance, feasibility, and usefulness~\citep{afzal2025beyond}. Moreover, useful creative exploration requires more than simply injecting randomness~\citep{schmidhuber2010formal}. Our work builds on these ideas by learning both \emph{how creatively} individual actions should be performed and \emph{when} different levels of creativity are useful across a reasoning trajectory.

\section{\textbf{\textsc{AI Night-Scientist}}: A Creativity-Aligned Agentic Framework}

\begin{figure}[h]
    \centering
    \includegraphics[width=1.0\linewidth]{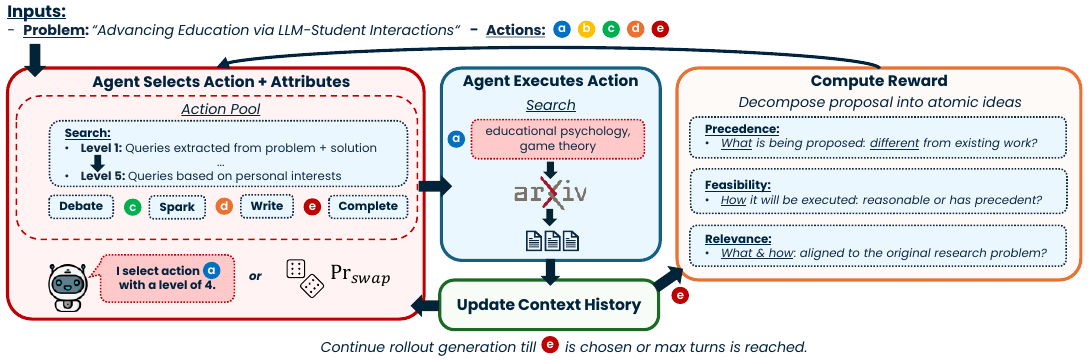}
    \caption{\textsc{AI Night-Scientist} consists of a: \textbf{(1)} rollout phase, where an LLM builds a reasoning trajectory $f$ by iteratively selecting and executing actions with creativity levels, and \textbf{(2)} reward phase during training, where the reward is computed over the final proposal.}
    \label{fig:framework}
\end{figure}

\par We propose \textbf{\textsc{AI Night-Scientist}}, as illustrated in Figure \ref{fig:framework}, which represents creativity directly within the agent's reasoning trajectory rather than only in its final output. We extend the ReAct-style reasoning setting~\citep{yao2023react}: at each step, the agent selects an action together with a creativity level that specifies \emph{how} that action should be carried out, executes it, and updates its current idea state. Repeating this process produces a trajectory through the space of possible ideas, allowing the agent to move between more familiar and more unexplored directions as reasoning unfolds. This lets us represent creativity at three levels (Figure \ref{fig:three_levels}): \textit{action-level} creativity captures \emph{how} an action is performed, \textit{process-level} creativity captures \emph{when} to shift between lower- and higher-creativity actions, and \textit{outcome-level} creativity captures the novelty and usefulness of the resulting idea.

\subsection{Multi-Level Representation of Creative Reasoning}
\label{sec:multi-level-representation}

\par We define each level below before applying the framework to scientific research proposal generation:

\begin{definition}[\textit{Action-Level Creativity}]
Let $\mathcal{A}$ denote the action space for a task $\mathcal{T}$. For each action $a \in \mathcal{A}$, we define an ordered set of creativity levels
$\mathcal{C}_a = \{c_a^{(1)}, \dots, c_a^{(K_a)}\}$,
where each level gives a natural-language description of how $a$ should be performed. Lower levels describe more conventional, high-probability behavior, while higher levels describe increasingly exploratory or unconventional behavior. The number of levels $K_a$ may vary across actions.
\end{definition}

\begin{definition}[\textit{Process-Level Creativity}]
Let $\tau = \bigl((a_1,c_1,\hat{o}_1), \dots, (a_N,c_N,\hat{o}_N)\bigr)$ denote a reasoning trajectory, where $a_i \in \mathcal{A}$ is the action selected at step $i$, $c_i \in \mathcal{C}_{a_i}$ is its creativity level, and $\hat{o}_i$ is the resulting intermediate output. At each step, the agent selects the next action-level choice $(a_i,c_i)$ based on the task $\mathcal{T}$ and the preceding trajectory $\tau_{1:i-1}$. \textit{Process-level creativity} reflects how effectively the agent sequences and adapts these choices over time. Higher process-level creativity means varying the degree of action-level creativity to benefit the evolving reasoning process, rather than consistently favoring either low- or high-creativity behavior.
\end{definition}

\begin{definition}[\textit{Outcome-Level Creativity}]
Let $o$ denote the final outcome produced for a task $\mathcal{T}$. \textit{Outcome-level creativity} captures the creativity of $o$ itself, based on two complementary properties: its \textit{novelty} $\mathcal{N}(o)$ and its \textit{usefulness} $\mathcal{U}(o)$~\citep{harvey2023toward}. Novelty measures how much $o$ departs from existing or familiar solutions, while usefulness measures how valuable, appropriate, or effective it is for the task. An outcome should exhibit both in order to be considered creative.
\end{definition}

\par Together, this multi-level representation separates \emph{how} creativity is expressed within an action, \emph{when} different degrees of creativity are useful across reasoning, and \emph{what} the process ultimately produces. We represent action-level creativity in natural language to make these choices interpretable and give the user direct semantic control over how and to what extent each action may deviate from its conventional execution. The number and meaning of creativity levels can also vary by action.

\par These action-level choices accumulate into the reasoning trajectory. Intermediate outputs may not appear directly in the final result, but they can shape later decisions and influence when greater or lesser deviation is useful. Greater creativity may help open new directions, while lower-creativity behavior may be better suited for developing or refining promising ones. Making these choices well requires the model to monitor how its reasoning is progressing and adapt accordingly, a key aspect of metacognitive control that is lacking in existing models \citep{kargupta2025cognitive}.

\subsection{Task Formulation: Research Proposal Generation}
\label{sec:task-formulation}

\par Given a high-level research problem $p$, the agent uses the action space $\mathcal{A}$ in Table~\ref{tab:tasks} to explore and develop a long-horizon research proposal over a multi-step trajectory $\tau$. \textsc{Search}, \textsc{Debate}, and \textsc{Spark} support multiple creativity levels, allowing the agent to vary how broadly it searches, whose perspectives it considers, and how strongly it challenges existing assumptions. \textsc{Write} consolidates the trajectory into proposal text, while \textsc{Stop} ends the process and returns the final proposal $o$. We keep \textsc{Write} fixed so that the final proposal primarily reflects the exploration that preceded it.

\begin{table*}[h!]
\scriptsize

\renewcommand{\arraystretch}{1.25}
\nstablerules

\renewcommand{\tabularxcolumn}[1]{>{\raggedright\arraybackslash}m{#1}}

\begin{tabularx}{\textwidth}{
    |>{\raggedright\arraybackslash}m{1.04cm}
    |>{\raggedright\arraybackslash}m{3.75cm}
    |>{\raggedright\arraybackslash\setlength{\parskip}{1.5pt}}X|
}
\hline

\rowcolor{ActionBlue}
\color{white}\textbf{Action $\mathcal{A}$}
&
\color{white}\textbf{Mechanism \& Output $\hat{o}$}
&
\color{white}\textbf{Creativity Levels $c \in \mathcal{C}_a$}
\\
\hline

\hyperref[app:search-action-prompt]
{\actioncircle{ActionBlue}{a}~Search}
&
Generate query $\rightarrow$ retrieve arXiv papers based on embedding similarity
&
\textit{L1:} Search proposal-specific background using title and core terms.

\textit{L3:} Search tangentially related background for broader context.

\textit{L5:} Search distant domains, alternate perspectives, or broader questions.
\\
\hline

\cellcolor{NSTableStripe}
\hyperref[app:debate-action-prompt]
{\actioncircle{ActionYellow}{b}~Debate}
&
\cellcolor{NSTableStripe}
Select participants and topic $\rightarrow$ retrieve relevant papers $\rightarrow$ simulate discussion
&
\cellcolor{NSTableStripe}
\textit{L1:} Discuss proposal specifics with a close-domain colleague.

\textit{L3:} Debate with a peer from the same field but a different topic.

\textit{L5:} Explore with experts from distant disciplines in an open-ended debate.
\\
\hline

\hyperref[app:spark-action-prompt]
{\actioncircle{ActionGreen}{c}~Spark}
&
Identify assumption (\textit{Bit}) $\rightarrow$ invert it (\textit{Flip}) $\rightarrow$ reframe it (\textit{Spark})
&
\textit{L1:} Challenge a narrow assumption specific to the current proposal.

\textit{L3:} Challenge a meaningful assumption underlying the approach.

\textit{L5:} Challenge a broad, field-level assumption through radical reframing.
\\
\hline

\cellcolor{NSTableStripe}
\hyperref[app:write-action-prompt]
{\actioncircle{ActionOrange}{d}~Write}
&
\cellcolor{NSTableStripe}
Synthesize prior trajectory into a proposal draft or revision
&
\cellcolor{NSTableStripe}
\textit{Fixed:} consolidate prior trajectory; preserve credit assignment
\\
\hline

\hyperref[app:action-prompts]
{\actioncircle{ActionRed}{e}~Stop}
&
End trajectory \& return final proposal
&
Judge final proposal by $\mathcal{P}$ (precedence), $\mathcal{F}$ (feasibility), and $\mathcal{R}$ (relevance)
\\
\hline

\end{tabularx}

\normaltablerules
\caption{Action space for research proposal generation with low, mid and high creativity levels shown. Table~\ref{tab:action-levels} in Appendix \ref{app:action-space} includes all levels; full action prompts provided in Appendix~\ref{app:action-prompts}.}
\label{tab:tasks}
\end{table*}

\par We focus on proposals with a scope similar to long-term research grants, where ideas are intended to guide work over several years rather than describe an immediately executable experiment. This makes the setting well suited to studying creative reasoning: proposals must remain grounded in existing work, yet many of their central ideas cannot be directly verified at inference time because the required experiments or data may not yet exist. The task therefore rewards reasoning that can move beyond established directions while still producing ideas that are feasible and relevant to $p$.

\subsection{Reinforcing Adaptive Creativity}

\par To \textit{teach} models \emph{when} and \emph{how} different degrees of creativity are useful, we leverage reinforcement learning. RL allows the quality of the final idea to shape the reasoning process without prescribing how that process should unfold. Our training pipeline consists of two phases: (i) a \textbf{rollout phase}, where the agent constructs a reasoning trajectory $\tau$ using the creativity-aligned action space above, and (ii) a \textbf{reward phase}, where the resulting proposal $o$ is evaluated. We use Group Relative Policy Optimization (GRPO)~\citep{shao2024deepseekmath}, which learns from relative rewards across sampled trajectories without requiring a separate critic. In our primary setting, the reward is applied only to the final proposal, requiring the model to learn which actions to take, how creatively to perform them, and how to sequence them based on their eventual effect on proposal quality.

\subsubsection{Rewarding Proposal Quality}
\label{sec:outcome-reward}

\par Scientific proposals contain multiple ideas whose novelty and feasibility may differ substantially, making a single holistic quality judgment difficult to ground. We therefore first decompose the final proposal into its atomic research ideas, separating \emph{what} each phase proposes from \emph{how} it plans to execute it. We then evaluate each component against the original research problem and relevant literature along three dimensions: \textit{precedence}, \textit{feasibility}, and \textit{relevance} (Table~\ref{tab:rewards}).

\begin{table}[h]
\scriptsize
\label{tab:rewards}
\renewcommand{\arraystretch}{1.3}
\nstablerules
\rowcolors{2}{NSTableStripe}{white}
\begin{tabularx}{\textwidth}{|p{1.1cm}|X|X|X|}
\hline
\rowcolor{ActionBlue}
\color{white}\textbf{Dimension} & \color{white}\textbf{Definition} & \color{white}\textbf{Positive Example (Reward $\uparrow$)} & \color{white}\textbf{Negative Example (Reward $\downarrow$)} \\
\hline
\hyperref[app:novelty-prompt]{\textbf{Precedence}} & Distance from prior work and existing approaches. & Core idea opens genuinely new technical directions. & Recombines familiar components without new insight. \\
\hline
\hyperref[app:feasibility-prompt]{\textbf{Feasibility}} & Credibility \& specificity of the execution plan. & Clear methods with precedent or scoped evaluation strategy. & Broad, underspecified plan or implausible scope. \\
\hline
\hyperref[app:relevance-prompt]{\textbf{Relevance}} & Alignment of the proposal with problem $p$. & Addresses a critical bottleneck in the target domain. & Drifts off-topic or fails to engage with the core challenges of $p$. \\
\hline
\end{tabularx}
\rowcolors{2}{}{}
\normaltablerules
\caption{Reward dimensions for proposal generation with examples. Reward prompts in Appendix~\ref{app:rewards}.}
\end{table}


\par We average precedence and relevance across the proposal's atomic ideas, but take the minimum feasibility across experimental plans, since a single infeasible phase may compromise the executability of the overall proposal. The resulting three proposal-level scores are then weighted equally to form the outcome reward: $R_{\mathrm{out}} = \frac{1}{3}
\left(R_{\mathcal{P}} + R_{\mathcal{F}} + R_{\mathcal{R}}\right)$. These rewards are designed to provide targeted training signals for proposal quality, with atomic ideas evaluated separately and precedence and feasibility grounded in retrieved literature.

\par Outcome-only training still leaves the model to discover useful creative behaviors through their eventual effect on proposal quality. We therefore explore two additional forms of guidance: exposing the model to occasional serendipitous actions early in training, and directly rewarding properties of the intermediate reasoning process. Neither is required by the framework; we study whether either provides additional benefit beyond the final outcome reward.

\subsubsection{Incentivizing Exploration through Serendipitous Actions}

\par Human discovery is often shaped by unexpected encounters that expose researchers to directions they may not have deliberately pursued~\citep{lubart2001models}. Similarly, an LLM early in training may repeatedly select familiar, low-creativity behaviors simply because it has not experienced useful alternatives. We therefore \emph{explore} a stochastic intervention that occasionally exposes the model to a different action-level choice. With probability $\Pr(\textit{swap})$, the selected pair $(a_i,c_i)$ is replaced by a sampled alternative $(a'_i,c'_i)$, and a decay factor $\gamma$ gradually reduces this probability over training. The goal is not to prescribe random exploration, but to expose the model early on to creative behaviors whose value it can later learn from the resulting reward.

\subsubsection{Exploring Process-Level Rewards}
\par Our primary models receive only the outcome reward above, allowing process-level behavior to emerge from credit assigned to the final proposal. We additionally investigate whether directly rewarding intermediate reasoning provides further benefit. The process reward evaluates each intermediate action along two complementary dimensions. \textit{Exploration} measures whether $a_i$ introduces a new direction relative to $\tau_{1:i-1}$, while \textit{contribution} measures how much $a_i$ ultimately contributes to the final proposal $o$. We score both dimensions for each intermediate action and average them across the trajectory to obtain $R_{\mathrm{proc}}$, which then forms the final reward $R_{\text{po}} = \frac{1}{2}\left(R_{\mathrm{proc}} + R_{\mathrm{out}}\right)$. Full details and prompts are provided in Appendix~\ref{app:process-reward}.

\section{Experiments}

\par We train all models using \texttt{verl} \citep{sheng2024hybridflow} with GRPO \citep{shao2024deepseekmath} on \texttt{Qwen3-8B-Base} and \texttt{Qwen3-14B-Base}. Each agent trajectory consists of up to 5 actions. For the \textit{search} action, we index arXiv\footnote{\texttt{https://arxiv.org/}} and use embedding-based retrieval with \texttt{GPT-4.1} as an external judge within the reward pipeline. Full training details are provided in Appendix~\ref{app:training-details}.

\subsection{Dataset}

\par We construct our dataset from the NSF Awards Database\footnote{https://huggingface.co/datasets/davidheineman/nsf-awards}, focusing on Computer Science, Engineering, and Mathematics (CSE) awards from 2018 onward. We use a $90{:}10$ train/test split, yielding 4,414 training and 491 test awards. Our model receives only the award title as $p$, which typically describes a broad, long-horizon research problem; we withhold the award abstract, project outcomes, and associated publications so that generation is not anchored to the funded solution. We focus on CSE to ensure broad coverage in open-access arXiv literature, while retaining substantial interdisciplinary breadth: the training set spans 46 research domains, with 23.1\% of subfield labels in core AI/ML and over 48\% of label occurrences outside AI/ML and core CS and Engineering (Appendix~\ref{app:dataset-details}).

\par Original NSF grant proposals are not publicly available, so we construct \emph{reconstructed reference proposals} from evidence surrounding each funded project. For each award, we collect its abstract, project outcomes report, PI information, and award period, then retrieve likely associated arXiv papers by the award PIs. Candidate papers are ranked using PI overlap, topical similarity, publication timing, and explicit funding acknowledgments. We prompt \texttt{GPT-5.1} to treat these sources as downstream evidence and reconstruct a plausible pre-award research plan that could have led to the observed project and publications. The reference follows the same structured format as generated proposals, including a summary, background, and multi-phase research plan.

\par These reconstructions are not intended to reproduce the original NSF proposals. Rather, they provide standardized, evidence-grounded references for research directions that were actually funded and subsequently pursued. We use them only as matched references for pairwise evaluation, and the proposal-generating model never receives the oracle information used to construct them. Full reconstruction details and prompts are provided in Appendix~\ref{app:reconstruction}.

\subsection{Evaluation Metrics}
\label{sec:eval_metrics}
\par Scientific creativity is difficult to capture with any single automatic metric, particularly because the long-term value of a research idea may not be observable for years. We therefore evaluate models along three complementary dimensions: \textit{predicted citation impact}, \textit{literature-grounded originality}, and \textit{research idea diversity}. Together, these capture both the quality of individual proposals and the range of research directions explored across the dataset.

\par \textbf{Predicted Citation Impact.}
For each problem, we compare the generated proposal against its matched reconstructed reference in randomized order and report pairwise win rate. We use \textbf{\texttt{SciJudge}}~\citep{tong2026ai}, a 30B-parameter model trained to predict relative citation impact from large-scale community signals, as a proxy for the potential usefulness and downstream value of the proposed research (prompts in Appendix~\ref{app:scijudge}). It has an average $82.7\%$ citation prediction accuracy.

\par \textbf{Literature-Grounded Originality.}
We compare each generated proposal against its matched reference for originality. To ground the judgment in prior work, we retrieve the closest paper to each proposal and ask \texttt{GPT-5.1} to make a randomized pairwise comparison relative to the retrieved literature (Appendix~\ref{app:llm-novelty}).

\par \textbf{Research Idea Diversity.}
Pairwise metrics capture the quality of individual proposals but not whether a model repeatedly produces the same kinds of ideas. Following the annotation format of \citet{chen2026measuring}, we use \texttt{GPT-5.4-mini} to classify each proposal by its primary \textit{research idea paradigm} and, analogously, by its primary \textit{contribution type} using categories adapted from prior taxonomies~\citep{wobbrock2012seven,miles2017taxonomy}. For each taxonomy, we report the effective number of categories represented (Appendix~\ref{app:contribution-type}), normalized by the seven available categories; higher values indicate a broader and more balanced range of research directions. The research-paradigm judge \citet{chen2026measuring} achieves $\kappa=0.84$ agreement with human judgments on 150 samples.

\par \textbf{Human-LLM Agreement.}
As an additional check on our pairwise judges, two human annotators evaluated a subset of proposals. Inter-annotator agreement was 80.0\% ($\kappa=0.625$) for impact and 86.7\% ($\kappa=0.766$) for originality, while human-LLM agreement was 76.7\% for \texttt{SciJudge} on impact and 72.4\% for \texttt{GPT-5.1} on originality (Appendix~\ref{app:human-eval}).

\subsection{Baselines}

\par We compare against baselines that vary how creativity is introduced into proposal generation. \textbf{Zero-shot} models directly generate a structured proposal from the input problem without retrieval, action selection, or creativity levels. \textbf{Temperature} retains the same five creativity levels, but maps each level only to decoding temperature, with $c\in\{1,\ldots,5\}$ corresponding to $T\in\{0,0.25,0.5,0.75,1.0\}$. This tests whether increased stochasticity alone can reproduce the benefits of semantic creativity guidance. \textbf{ReAct} \citep{yao2023react} uses the same action space $\mathcal{A}$ with no corresponding $\mathcal{C}_a$.

\par \textbf{Creative} baselines apply our full creativity-aligned action space and natural-language creativity levels at inference time, but receive no RL training. We also train several \textbf{RL variants} with the same outcome-level reward as \textsc{AI Night-Scientist}: \textbf{+ Temp} and \textbf{+ ReAct} apply RL to the corresponding baselines above, while \textbf{+ Search + Write} retains semantic creativity levels but restricts $\mathcal{A}$ to retrieval and writing, similar to retrieval-augmented ideation systems~\citep{chenglillmideas, wang2024scimon}. Together, these separate the effects of semantic creativity guidance, RL, and the broader action space.

\par Finally, we compare against \textbf{GIANTS}~\citep{he2026giants}, an RL-trained model for scientific insight anticipation. Because GIANTS expects two input papers, we retrieve the two most relevant arXiv papers using the award title and abstract and reformat them with \texttt{GPT-4.1}. GIANTS therefore receives more award-specific information than \textsc{AI Night-Scientist}, which sees only the title.

\par For closed-source baselines, we use \texttt{GPT-4.1}, a strong general-purpose, non-reasoning model. Model recency does not necessarily imply greater creative diversity, with recent work finding increasing similarity across generations on open-ended tasks~\citep{patel2026llms}.

\subsection{Experimental Results \& Analysis}
\begin{table*}[h]
\centering
\tiny

\begin{minipage}[t]{0.625\textwidth}
\centering
\textbf{(a) Proposal quality}

\vspace{2pt}

\renewcommand{\arraystretch}{1.1}
\setlength{\tabcolsep}{3pt}
\resulttablerules

\begin{tabular}{|p{1cm}|l|c|c|}
\hline
\rowcolor{ResultHeader}
\textbf{Category} & \textbf{Method}
& \textbf{Citation (\%)} $\uparrow$
& \textbf{Originality (\%)} $\uparrow$ \\
\hline

\multirow{4}{=}{Zero-shot}
  & Llama-3.1-8B & 1.61 & 1.15 \\
  & \cellcolor{ResultAlt}Qwen3-8B
  & \cellcolor{ResultAlt}0.46
  & \cellcolor{ResultAlt}2.53 \\
  & Qwen3-14B & 1.15 & 2.53 \\
  & \cellcolor{ResultAlt}GPT-4.1
  & \cellcolor{ResultAlt}3.90
  & \cellcolor{ResultAlt}12.18 \\
\hline

\multirow{2}{=}{Temp.}
  & Qwen3-8B + Temp & 2.07 & 0.69 \\
  & \cellcolor{ResultAlt}GPT-4.1 + Temp
  & \cellcolor{ResultAlt}2.99
  & \cellcolor{ResultAlt}9.89 \\
\hline

\multirow{2}{=}{ReAct}
  & Qwen3-8B + ReAct & 2.33 & 6.54 \\
  & \cellcolor{ResultAlt}GPT-4.1 + ReAct
  & \cellcolor{ResultAlt}3.45
  & \cellcolor{ResultAlt}10.11 \\
\hline

\multirow{4}{=}{Creative}
  & Llama-3.1-8B + Creative & 1.10 & 6.52 \\
  & \cellcolor{ResultAlt}Qwen3-8B + Creative
  & \cellcolor{ResultAlt}1.89
  & \cellcolor{ResultAlt}4.25 \\
  & Qwen3-14B + Creative & 0.71 & 4.76 \\
  & \cellcolor{ResultAlt}GPT-4.1 + Creative
  & \cellcolor{ResultAlt}3.00
  & \cellcolor{ResultAlt}14.02 \\
\hline

\multirow{4}{=}{w/ RL}
  & GIANTS-4B & 1.62 & 35.57 \\
  & \cellcolor{ResultAlt}Qwen3-8B + Temp
  & \cellcolor{ResultAlt}11.52
  & \cellcolor{ResultAlt}19.82 \\
  & Qwen3-8B + ReAct & 24.94 & 46.60 \\
  & \cellcolor{ResultAlt}Qwen3-8B + Search + Write
  & \cellcolor{ResultAlt}24.47
  & \cellcolor{ResultAlt}33.33 \\
\hline

\multirow{2}{=}{Ours}
  & \textbf{\textsc{AI Night-Scientist-8B}}
  & \underline{29.89}
  & \underline{56.32} \\
  & \cellcolor{ResultAlt}\textbf{\textsc{AI Night-Scientist-14B}}
  & \cellcolor{ResultAlt}\textbf{33.18}
  & \cellcolor{ResultAlt}\textbf{68.68} \\
\hline
\end{tabular}

\normaltablerules
\end{minipage}
\hfill
\begin{minipage}[t]{0.35\textwidth}
\centering

\textbf{(b) Research-idea diversity}

\vspace{2pt}

\renewcommand{\arraystretch}{1.18}
\setlength{\tabcolsep}{4pt}
\resulttablerules

\begin{tabular}{|l|c|c|}
\hline
\rowcolor{ResultHeader}
\textbf{Method}
& \textbf{Paradigm} $\uparrow$
& \textbf{Contri.} $\uparrow$ \\
\hline

Qwen3-8B Zero-Shot & 0.738 & 0.370 \\

\rowcolor{ResultAlt}
GPT-4.1 Zero-Shot
& 0.785
& \textbf{0.439} \\

\hline

GIANTS-4B (RL)
& 0.773
& 0.347 \\

\rowcolor{ResultAlt}
Qwen3 + Temp (RL)
& 0.792
& 0.339 \\

Qwen3 + ReAct (RL)
& \underline{0.888}
& 0.390 \\

\hline

\rowcolor{ResultAlt}
Night-8B ($R_{\text{po}}$)
& 0.853
& 0.362 \\

+ No swap
& 0.844
& 0.358 \\

\rowcolor{ResultAlt}
\textbf{Night-8B ($R_{\text{out}}$)}
& \textbf{0.943}
& \underline{0.425} \\

\hline
\end{tabular}

\normaltablerules
\vspace{6pt}
\textbf{(c) Reward ablation}

\vspace{2pt}

\renewcommand{\arraystretch}{1.15}
\setlength{\tabcolsep}{4pt}
\resulttablerules

\begin{tabular}{|l|l|c|c|}
\hline
\rowcolor{ResultHeader}
\textbf{Method}
& \textbf{Reward}
& \textbf{Citation} $\uparrow$
& \textbf{Original.} $\uparrow$ \\
\hline

\multirow{2}{*}{ReAct}
& $R_{\text{out}}$ & 24.94 & 46.60 \\
& $R_{\text{po}}$ & 2.10 & 6.54 \\
\hline

\multirow{2}{*}{\cellcolor{ResultAlt}Temp.}
& \cellcolor{ResultAlt}$R_{\text{out}}$
& \cellcolor{ResultAlt}11/52
& \cellcolor{ResultAlt}19.82 \\
&
\cellcolor{ResultAlt}$R_{\text{po}}$
& \cellcolor{ResultAlt}10.35
& \cellcolor{ResultAlt}13.33 \\
\hline

\multirow{2}{*}{Night-8B}
& $R_{\text{out}}$
& \textbf{29.89}
& \underline{56.32} \\
& $R_{\text{po}}$
& \underline{26.20}
& \textbf{61.61} \\
\hline

\end{tabular}

\normaltablerules
\end{minipage}

\caption{
Main results for \textbf{(a)} proposal quality, \textbf{(b)} research-idea diversity, and \textbf{(c)} outcome-only ($R_{\text{out}}$) vs.\ process + outcome ($R_{\text{po}}$) training. Best results are bolded and second-best are underlined.
}
\label{tab:main_results}
\vspace{-5mm}
\end{table*}

\subsubsection{Creativity-Aligned RL Improves Proposal Quality}

\par Table~\ref{tab:main_results}(a) shows a consistent advantage for \textsc{AI Night-Scientist} over both direct-generation and agentic baselines. Relative to \texttt{Qwen3-8B}, \textsc{Night-8B} improves predicted citation impact by 29.43 percentage points and originality by 53.79 points. At 14B, these gains grow to 32.03 and 66.15 points over Qwen3-14B. In comparison, scaling the zero-shot model from \texttt{8B} to \texttt{14B} yields almost no improvement, whereas scaling \textsc{AI Night-Scientist} adds another 3.29 points in citation and 12.36 points in originality. \textit{Additional model capacity therefore appears substantially more useful once paired with a learned creative reasoning policy.}

\par The contrast with temperature-based exploration is especially evident after RL. Although both optimize the same proposal-level reward, \textsc{Night-8B} exceeds the temperature-controlled variant by 36.50 points in originality and 18.37 points in citation. ReAct closes much of this gap, showing that a learned multi-step policy already helps, but semantic creativity levels still add 9.72 points in originality and 4.95 points in citation. \textit{The useful signal is therefore not simply to ``explore more,'' but to specify how an action should deviate so RL can learn when those deviations are useful.}

\par The action space matters as well. Restricting the agent to \textit{search} and \textit{write} reduces originality by 22.99 points despite using the same creativity levels and outcome reward. Retrieval and iterative writing therefore explain only part of the gain; \textit{spark} and \textit{debate} provide additional ways to redirect reasoning rather than simply gather more evidence for an existing direction.

\paragraph{\textbf{Robustness.}} To verify that these trends are not specific to reconstructed references, we directly compare \textsc{Night-8B} against GPT-4.1 and \texttt{Qwen3-8B}. It wins 74.95\% of citation and 86.96\% of originality comparisons against GPT-4.1, increasing to 82.48\% and 97.56\% against \texttt{Qwen3-8B}. The same advantage therefore holds in direct head-to-head comparisons. We also include the confidence interval results under Appendix \ref{app:main-results-ci}.

\subsubsection{\textsc{AI Night-Scientist} Broadens Research Idea Diversity}

\par Higher originality would be less meaningful if the model repeatedly relied on the same kind of creative strategy. Table~\ref{tab:main_results}(b) suggests otherwise. Relative to zero-shot Qwen3-8B, \textsc{Night-8B} increases normalized research-paradigm coverage from 0.738 to 0.943, corresponding to roughly 1.4 additional effective categories out of seven; contribution-type coverage similarly rises from 0.370 to 0.425. \textit{The model therefore produces not only more original proposals, but a broader range of research paradigms and contribution types.}

\par Diversity is also sensitive to early exploration. Removing serendipitous action swaps reduces paradigm coverage by 0.099, or roughly 0.7 effective categories, and contribution coverage by 0.067, or roughly 0.5 categories. This pattern is consistent with early exposure to less familiar behaviors helping the policy discover a broader set of useful trajectories.

\par The improvements are also not concentrated in a small set of CS topics. \textsc{Night-8B} improves over zero-shot Qwen3-8B on both citation and originality across all 36 evaluated domains, including healthcare, biomedical engineering, geosciences, and quantum science. Full per-domain results are provided in Appendix~\ref{app:domain_gen}.

\subsubsection{Process Rewards Favor Originality over Breadth}

\par Table~\ref{tab:main_results}(c) shows a different effect from directly rewarding the reasoning process. For \textsc{Night-8B}, process + outcome training raises originality by 5.29 points but lowers predicted citation impact by 3.69 points. Paradigm and contribution coverage also fall by roughly 0.6 and 0.4 effective categories, respectively. Process supervision can therefore favor more original individual proposals without necessarily producing a broader or more impactful set of ideas.

\par This effect is specific to the creativity-aligned action space. Adding the same process reward to ReAct sharply reduces both citation and originality, while Temperature also declines on both metrics. Process supervision is therefore not uniformly beneficial; its effect depends on whether the action space provides semantically distinct ways to express different degrees of creativity. We consequently use outcome-only training as our primary setting and treat process rewards as a way to shift the model toward greater proposal-level originality. We detail further analysis of $P_{\texttt{po}}$ in Appendix \ref{app:process-actions}.

\subsubsection{Qualitative Analysis}

\par We examine one representative NSF topic to understand how proposal directions differ qualitatively across methods. Table~\ref{tab:qualitative_comparison} shows a progression from adapting existing lecture-delivery mechanisms toward changing the underlying learning interaction itself. \textsc{AI Night-Scientist} produces the more substantial reframings, although the process + outcome variant also illustrates that greater creative deviation can come at the cost of grounding and methodological rigor.

\begin{table}[h]
\scriptsize
\label{tab:qualitative_comparison}
\renewcommand{\arraystretch}{1.3}
\nstablerules
\rowcolors{2}{NSTableStripe}{white}
\begin{tabularx}{\textwidth}{|p{1.cm}|X|p{5.7cm}|}
\hline
\rowcolor{ActionBlue}
\color{white}\textbf{Method} &
\color{white}\textbf{Key Proposal Direction} &
\color{white}\textbf{Main Limitation} \\
\hline

GPT-4.1 &
Uses AI to \textcolor{red!70!black}{segment prerecorded lectures and represent the material through conversational agents}, with adaptive pacing, clarification, and multimodal accessibility support. &
The proposal is coherent and inclusive, but largely \textcolor{red!70!black}{combines established lecture-segmentation and chatbot mechanisms} rather than introducing a distinct technical idea. \\
\hline

ReAct &
Introduces a \textcolor{green!60!black}{multimodal engagement score that combines visual, auditory, and physiological signals} and uses it to adapt lecture pacing and content granularity in real time. &
The score is a concrete artifact, but the proposed \textcolor{red!70!black}{RL-based adaptation is underspecified}: the state, action, and reward spaces are left unclear for how engagement signals translate into adaptation decisions. \\
\hline

\textsc{Night-8B} ($R_{\text{out}}$) &
Proposes \textcolor{green!60!black}{\textit{Agentify}, which turns a prerecorded lecture into a live, agent-mediated session} where an AI interleaves questions, hints, and dialogue based on learner behavior. &
The proposal substantially reframes the interaction, but some components are \textcolor{red!70!black}{only loosely motivated or underdeveloped}, including the cognitive-tier co-evolution mechanism and forced speed controls. \\
\hline

\textsc{Night-8B} ($R_{\text{po}}$) &
Proposes \textcolor{green!60!black}{Progressive Content Enactors (PCEs), which deliberately inject controlled errors and ambiguities for learners to detect and resolve}, shifting the lecture from passive delivery toward active problem solving. &
The central idea is distinctive, but the proposal also contains \textcolor{red!70!black}{unverifiable quantitative claims and weakly grounded evaluation measures}, reducing methodological credibility. \\
\hline

\end{tabularx}
\rowcolors{2}{}{}
\normaltablerules
\caption{Qualitative comparison for the NSF topic \textit{``Using AI to Transform Online Video Lectures.''} 
Green highlights novel or well-specified contributions; red highlights incremental, underspecified, or questionable elements. 
Full proposal excerpts and detailed error analysis are provided in Appendix~\ref{app:qualitative_analysis}.}
\end{table}

\begin{wrapfigure}[10]{r}{0.45\textwidth}
    \vspace{-1\baselineskip}
    \centering
    \includegraphics[width=\linewidth]{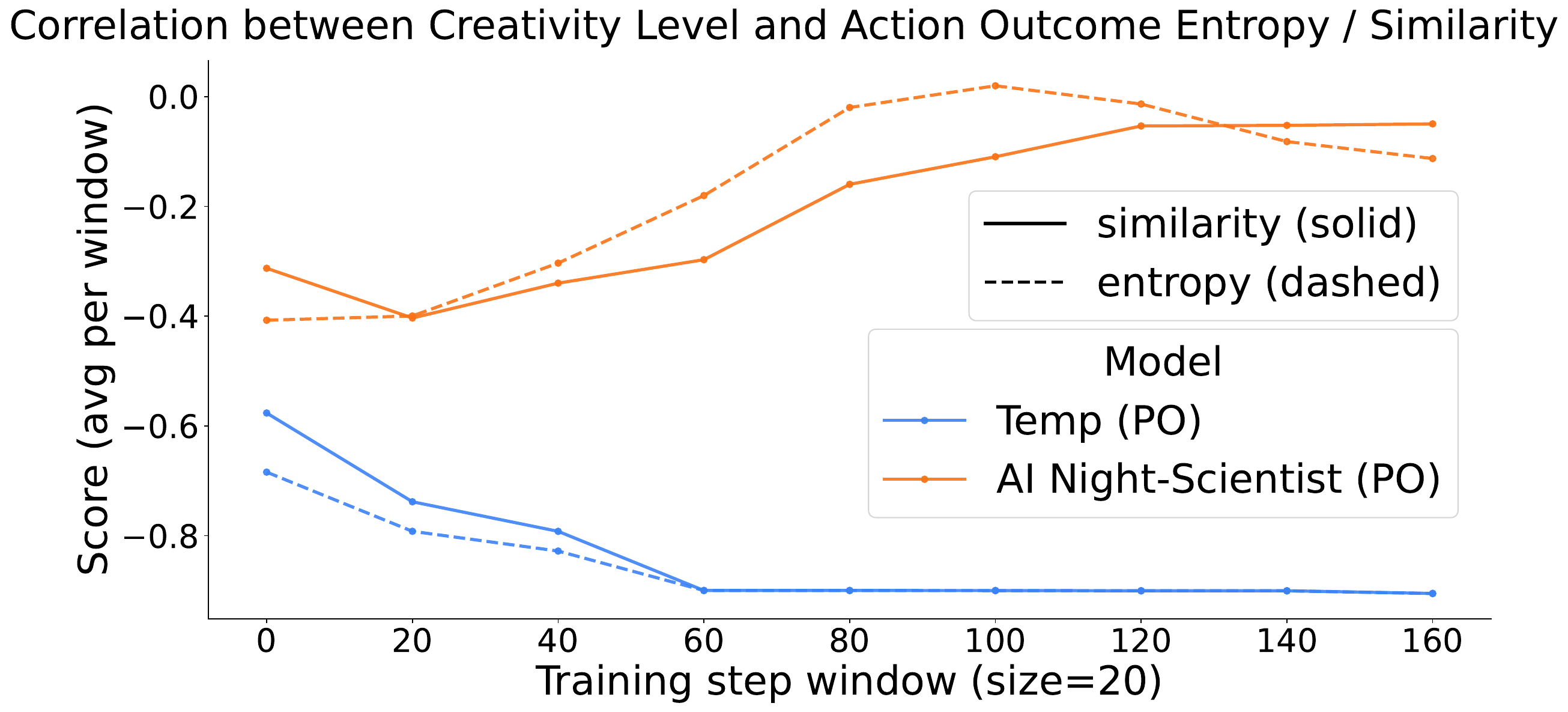}
    \caption{Higher correlation indicates better alignment between selected creativity level and output behavior.}
    \label{fig:action_level_entropy_similarity}
\end{wrapfigure}

\subsubsection{Semantic Creativity Levels Better Align with Output Behavior}

Our creativity levels are intended to change how an action is executed, not merely label it. Figure~\ref{fig:action_level_entropy_similarity} measures this relationship using output entropy and semantic similarity (Appendix~\ref{app:action-level-metrics}). \textsc{AI Night-Scientist~(PO)} improves on both measures during training, while Temp~(PO) deteriorates toward $-0.9$. \textit{Semantic descriptions therefore provide a more learnable link between the selected creativity level and resulting behavior than temperature alone.}
\subsubsection{Process Rewards Shift the Learned Action Policy}
\label{app:process-actions}

\begin{figure}[h]
    \centering
    \includegraphics[width=1\linewidth]{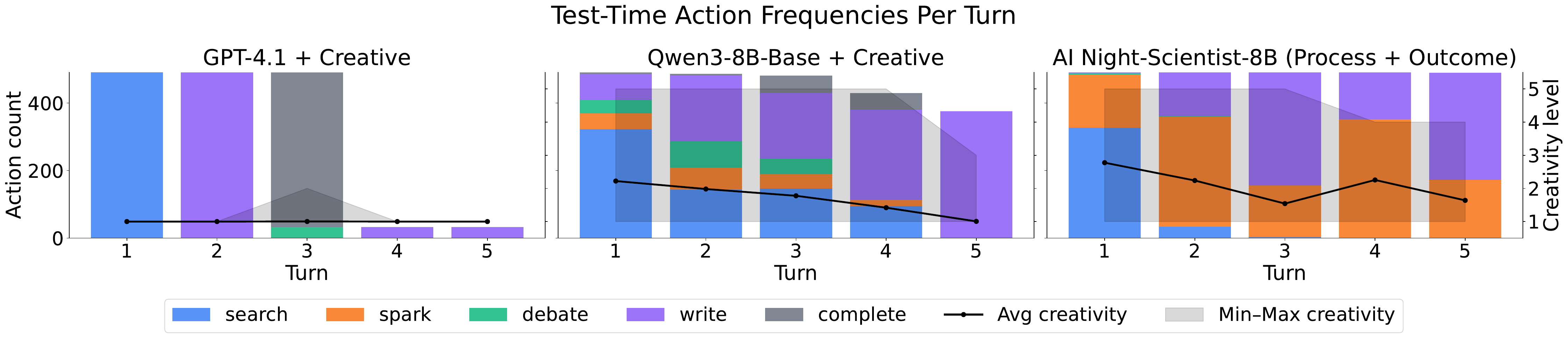}
    \caption{Test-time action frequencies across five reasoning turns for three model variants.}
    \label{fig:exploration_vs_contribution}
\end{figure}

\par Figure~\ref{fig:exploration_vs_contribution} shows how the learned policies differ at test time. \texttt{GPT-4.1} typically follows a short \textit{search}~$\rightarrow$~\textit{write}~$\rightarrow$~\textit{stop} pattern, while \texttt{Qwen3-8B-Base} uses a broader mix of actions across the full trajectory. \textsc{AI Night-Scientist-8B} trained with process+outcome rewards shows a different preference: it selects \textit{spark} more frequently across turns and tends to use it at moderate-to-high creativity levels. This suggests that process supervision shifts the policy toward assumption-challenging actions that more directly open new research directions.

\par \textit{Debate} is selected relatively rarely. One practical reason may be that it requires more of the trajectory budget, since the agent first identifies participants and relevant evidence before generating the grounded discussion. In general, this suggests that the learned value of an action depends not only on its creative potential, but also on how efficiently it contributes within a limited reasoning horizon.

\section{Conclusion}

\par We introduce \textsc{AI Night-Scientist}, a creativity-aligned agentic framework that represents creativity at the action, process, and outcome levels and uses reinforcement learning to teach models \emph{when} and \emph{how} to depart from predictable reasoning. Applied to long-horizon research proposal generation, \textsc{AI Night-Scientist} improves predicted citation impact and originality by up to 32.03 and 66.15 points over its base model, while producing a broader range of research paradigms. These gains are not reproduced by higher decoding temperature or ReAct alone, supporting our central finding that creativity is more than sampling stochasticity. Our ablations further show that outcome-only training provides the strongest overall balance of quality and diversity, while process-level rewards can increase originality at the cost of citation impact and breadth. Together, these results suggest that creativity can be learned as a multi-level reasoning capability, enabling scientific agents to move more flexibly between the structured reasoning of \textit{day science} and the exploratory reasoning of \textit{night science} while remaining tools for human-led discovery.

\subsection*{Ethics Statement}

\paragraph{Supporting human-led research.}
\textsc{AI Night-Scientist} is designed for early-stage scientific ideation, when research directions are still speculative and not yet fully testable. Its role is to help researchers explore a broader set of possible directions, including connections or assumptions they may not otherwise consider. The system is intended as a creative collaborator rather than an autonomous researcher: generated proposals should be treated as candidate ideas that require critical evaluation, refinement, and validation by domain experts before they can support scientific conclusions.

\paragraph{Broadening research exploration.}
Human ideation is often shaped by the concepts and examples already available in their local research community~\citep{lubart2001models}. By encouraging proposals that depart from existing work while remaining relevant and feasible, \textsc{AI Night-Scientist} aims to surface directions beyond those most immediately accessible to a researcher or model. This may be especially useful for interdisciplinary exploration, where relevant ideas can be distributed across distant literature and research communities. Importantly, broader exploration does not imply that unfamiliar ideas are inherently better; their scientific value must still be established through expert judgment and empirical validation.

\paragraph{Risks and responsible use.}
As with other generative systems for scientific writing, \textsc{AI Night-Scientist} can produce plausible but incorrect, infeasible, or insufficiently grounded proposals, and could be misused to generate low-quality scientific content at scale. Its outputs should therefore be presented as AI-generated suggestions rather than validated research plans. Responsible use requires transparent disclosure, independent verification of claims and citations, and meaningful human oversight before ideas are pursued, disseminated, or incorporated into scientific work.

\subsection*{Reproducibility Statement}

\par The main paper specifies the action space, creativity levels, reward formulation, baselines, and evaluation metrics. The appendix further provides the complete action and reward prompts (Appendices~\ref{app:rewards}-\ref{app:action-prompts}), training compute and hyperparameters (Appendix~\ref{app:training-details}), NSF dataset filtering and domain construction (Appendix~\ref{app:dataset-details}), reconstructed reference-proposal procedure and prompt (Appendix~\ref{app:reconstruction}), and full evaluation protocols for predicted citation impact, originality, and research-idea diversity (Appendices~\ref{app:scijudge}-\ref{app:llm-novelty}). We additionally report confidence intervals and human-LLM agreement in Appendices~\ref{app:main-results-ci} and~\ref{app:human-eval}. Together, these materials document the data construction, model training, prompting, retrieval, and evaluation procedures needed to reproduce the reported experiments.

\bibliography{iclr2027_conference}

@article{dwyer2025evaluation,
  title={An Evaluation of the Relationship Between Critical Thinking and Creative Thinking: Complementary Metacognitive Processes or Strange Bedfellows?},
  author={Dwyer, Christopher P and Campbell, Deagl{\'a}n and Seery, Niall},
  journal={Journal of Intelligence},
  volume={13},
  number={2},
  pages={23},
  year={2025},
  publisher={MDPI}
}

@article{de2022reasoning,
  title={Reasoning outside the box: Divergent thinking is related to logical reasoning},
  author={de Chantal, Pier-Luc and Markovits, Henry},
  journal={Cognition},
  volume={224},
  pages={105064},
  year={2022},
  publisher={Elsevier}
}

@article{zheng2023judging,
  title={Judging llm-as-a-judge with mt-bench and chatbot arena},
  author={Zheng, Lianmin and Chiang, Wei-Lin and Sheng, Ying and Zhuang, Siyuan and Wu, Zhanghao and Zhuang, Yonghao and Lin, Zi and Li, Zhuohan and Li, Dacheng and Xing, Eric and others},
  journal={Advances in neural information processing systems},
  volume={36},
  pages={46595--46623},
  year={2023}
}

@inproceedings{zhong2024can,
  title={Can llm replace stack overflow? a study on robustness and reliability of large language model code generation},
  author={Zhong, Li and Wang, Zilong},
  booktitle={Proceedings of the AAAI conference on artificial intelligence},
  volume={38},
  number={19},
  pages={21841--21849},
  year={2024}
}

@article{liu2024exploring,
  title={Exploring and evaluating hallucinations in llm-powered code generation},
  author={Liu, Fang and Liu, Yang and Shi, Lin and Huang, Houkun and Wang, Ruifeng and Yang, Zhen and Zhang, Li and Li, Zhongqi and Ma, Yuchi},
  journal={arXiv preprint arXiv:2404.00971},
  year={2024}
}

@article{guo2025deepseek,
  title={Deepseek-r1: Incentivizing reasoning capability in llms via reinforcement learning},
  author={Guo, Daya and Yang, Dejian and Zhang, Haowei and Song, Junxiao and Zhang, Ruoyu and Xu, Runxin and Zhu, Qihao and Ma, Shirong and Wang, Peiyi and Bi, Xiao and others},
  journal={arXiv preprint arXiv:2501.12948},
  year={2025}
}

@article{wang2025reinforcement,
  title={Reinforcement learning for reasoning in large language models with one training example},
  author={Wang, Yiping and Yang, Qing and Zeng, Zhiyuan and Ren, Liliang and Liu, Liyuan and Peng, Baolin and Cheng, Hao and He, Xuehai and Wang, Kuan and Gao, Jianfeng and others},
  journal={arXiv preprint arXiv:2504.20571},
  year={2025}
}

@article{wechsler2018creative,
  title={Creative and critical thinking: Independent or overlapping components?},
  author={Wechsler, Solange Muglia and Saiz, Carlos and Rivas, Silvia F and Vendramini, Claudete Maria Medeiros and Almeida, Leandro S and Mundim, Maria Celia and Franco, Amanda},
  journal={Thinking skills and creativity},
  volume={27},
  pages={114--122},
  year={2018},
  publisher={Elsevier}
}

@article{halpern2007nature,
  title={The nature and nurture of critical thinking},
  author={Halpern, Diane F},
  journal={Critical thinking in psychology},
  number={1},
  pages={1--14},
  year={2007}
}

@article{lubart2001models,
  title={Models of the creative process: Past, present and future},
  author={Lubart, Todd I},
  journal={Creativity research journal},
  volume={13},
  number={3-4},
  pages={295--308},
  year={2001},
  publisher={Taylor \& Francis}
}

@article{harvey2023toward,
  title={Toward a meta-theory of creativity forms: How novelty and usefulness shape creativity},
  author={Harvey, Sarah and Berry, James W},
  journal={Academy of Management Review},
  volume={48},
  number={3},
  pages={504--529},
  year={2023},
  publisher={Academy of Management Briarcliff Manor, NY}
}

@article{cohen1989continuum,
  title={A continuum of adaptive creative behaviors},
  author={Cohen, Leonora M},
  journal={Creativity Research Journal},
  volume={2},
  number={3},
  pages={169--183},
  year={1989},
  publisher={Taylor \& Francis}
}

@article{mccoy2024embers,
  title={Embers of autoregression show how large language models are shaped by the problem they are trained to solve},
  author={McCoy, R Thomas and Yao, Shunyu and Friedman, Dan and Hardy, Mathew D and Griffiths, Thomas L},
  journal={Proceedings of the National Academy of Sciences},
  volume={121},
  number={41},
  pages={e2322420121},
  year={2024},
  publisher={National Academy of Sciences}
}

@article{agarwal2025unreasonable,
  title={The unreasonable effectiveness of entropy minimization in llm reasoning},
  author={Agarwal, Shivam and Zhang, Zimin and Yuan, Lifan and Han, Jiawei and Peng, Hao},
  journal={arXiv preprint arXiv:2505.15134},
  year={2025}
}

@article{zhao2025assessing,
  title={Assessing and understanding creativity in large language models},
  author={Zhao, Yunpu and Zhang, Rui and Li, Wenyi and Li, Ling},
  journal={Machine Intelligence Research},
  volume={22},
  number={3},
  pages={417--436},
  year={2025},
  publisher={Springer}
}

@article{he2026giants,
  title={GIANTS: Generative Insight Anticipation from Scientific Literature},
  author={He-Yueya, Joy and Singh, Anikait and Gao, Ge and Li, Michael Y and Yang, Sherry and Finn, Chelsea and Brunskill, Emma and Goodman, Noah D},
  journal={arXiv preprint arXiv:2604.09793},
  year={2026}
}

@article{gupta2025all,
  title={All that glitters is not novel: Plagiarism in ai generated research},
  author={Gupta, Tarun and Pruthi, Danish},
  journal={arXiv preprint arXiv:2502.16487},
  year={2025}
}

@article{wenger2025we,
  title={We're Different, We're the Same: Creative Homogeneity Across LLMs},
  author={Wenger, Emily and Kenett, Yoed},
  journal={arXiv preprint arXiv:2501.19361},
  year={2025}
}

@article{shao2024deepseekmath,
  title={Deepseekmath: Pushing the limits of mathematical reasoning in open language models},
  author={Shao, Zhihong and Wang, Peiyi and Zhu, Qihao and Xu, Runxin and Song, Junxiao and Bi, Xiao and Zhang, Haowei and Zhang, Mingchuan and Li, YK and Wu, Yang and others},
  journal={arXiv preprint arXiv:2402.03300},
  year={2024}
}

@article{afzal2025beyond,
  title={Beyond" Not Novel Enough": Enriching Scholarly Critique with LLM-Assisted Feedback},
  author={Afzal, Osama Mohammed and Nakov, Preslav and Hope, Tom and Gurevych, Iryna},
  journal={arXiv preprint arXiv:2508.10795},
  year={2025}
}

@inproceedings{wang2024scimon,
  title={Scimon: Scientific inspiration machines optimized for novelty},
  author={Wang, Qingyun and Downey, Doug and Ji, Heng and Hope, Tom},
  booktitle={Proceedings of the 62nd Annual Meeting of the Association for Computational Linguistics (Volume 1: Long Papers)},
  pages={279--299},
  year={2024}
}

@article{gu2024llms,
  title={LLMs can realize combinatorial creativity: generating creative ideas via LLMs for scientific research},
  author={Gu, Tianyang and Wang, Jingjin and Zhang, Zhihao and Li, HaoHong},
  journal={arXiv preprint arXiv:2412.14141},
  year={2024}
}

@article{yanai2019night,
  title={Night science},
  author={Yanai, Itai and Lercher, Martin},
  journal={Genome Biology},
  volume={20},
  number={1},
  pages={179},
  year={2019},
  publisher={Springer}
}

@article{gottweis2025towards,
  title={Towards an AI co-scientist},
  author={Gottweis, Juraj and Weng, Wei-Hung and Daryin, Alexander and Tu, Tao and Palepu, Anil and Sirkovic, Petar and Myaskovsky, Artiom and Weissenberger, Felix and Rong, Keran and Tanno, Ryutaro and others},
  journal={arXiv preprint arXiv:2502.18864},
  year={2025}
}

@inproceedings{yao2023react,
  title = {{ReAct}: Synergizing Reasoning and Acting in Language Models},
  author = {Yao, Shunyu and Zhao, Jeffrey and Yu, Dian and Du, Nan and Shafran, Izhak and Narasimhan, Karthik and Cao, Yuan},
  booktitle = {International Conference on Learning Representations (ICLR) },
  year = {2023},
  html = {https://arxiv.org/abs/2210.03629},
}

@article{tong2026ai,
  title={AI Can Learn Scientific Taste},
  author={Tong, Jingqi and Li, Mingzhe and Li, Hangcheng and Yang, Yongzhuo and Mou, Yurong and Ma, Weijie and Xi, Zhiheng and Chen, Hongji and Liu, Xiaoran and Cheng, Qinyuan and others},
  journal={arXiv preprint arXiv:2603.14473},
  year={2026}
}

@inproceedings{chenglillmideas,
 author = {Si, Chenglei and Yang, Diyi and Hashimoto, Tatsunori},
 booktitle = {International Conference on Learning Representations},
 editor = {Y. Yue and A. Garg and N. Peng and F. Sha and R. Yu},
 pages = {94003--94092},
 title = {Can LLMs Generate Novel Research Ideas? A Large-Scale Human Study with 100+ NLP Researchers},
 url = {https://proceedings.iclr.cc/paper_files/paper/2025/file/ea94957d81b1c1caf87ef5319fa6b467-Paper-Conference.pdf},
 volume = {2025},
 year = {2025}
}

@article{stent1988statue,
  title={The Statue Within: An Autobiography},
  author={Stent, Gunther S},
  journal={Science},
  volume={239},
  number={4847},
  pages={1545--1547},
  year={1988},
  publisher={American Association for the Advancement of Science}
}

@article{kargupta2025cognitive,
  title={Cognitive foundations for reasoning and their manifestation in llms},
  author={Kargupta, Priyanka and Li, Shuyue Stella and Wang, Haocheng and Lee, Jinu and Chen, Shan and Ahia, Orevaoghene and Light, Dean and Griffiths, Thomas L and Kleiman-Weiner, Max and Han, Jiawei and others},
  journal={arXiv preprint arXiv:2511.16660},
  year={2025}
}

@inproceedings{kargupta2025tree,
  title={Tree-of-debate: Multi-persona debate trees elicit critical thinking for scientific comparative analysis},
  author={Kargupta, Priyanka and Agarwal, Ishika and August, Tal and Han, Jiawei},
  booktitle={Proceedings of the 63rd Annual Meeting of the Association for Computational Linguistics (Volume 1: Long Papers)},
  pages={29378--29403},
  year={2025}
}

@article{kargupta2026sparking,
  title={Sparking Scientific Creativity via LLM-Driven Interdisciplinary Inspiration},
  author={Kargupta, Priyanka and Mehri, Shuhaib and Hakkani-Tur, Dilek and Han, Jiawei},
  journal={arXiv preprint arXiv:2603.12226},
  year={2026}
}

@inproceedings{kargupta2025beyond,
  title={Beyond true or false: Retrieval-augmented hierarchical analysis of nuanced claims},
  author={Kargupta, Priyanka and Tian, Runchu and Han, Jiawei},
  booktitle={Proceedings of the 63rd Annual Meeting of the Association for Computational Linguistics (Volume 1: Long Papers)},
  pages={29664--29679},
  year={2025}
}

@article{o2025sparks,
  title={Sparks of science: Hypothesis generation using structured paper data},
  author={O'Neill, Charles and Ghosal, Tirthankar and R{\u{a}}ileanu, Roberta and Walmsley, Mike and Bui, Thang and Schawinski, Kevin and Ciuc{\u{a}}, Ioana},
  journal={arXiv preprint arXiv:2504.12976},
  year={2025}
}

@article{lu2024llm,
  title={LLM discussion: Enhancing the creativity of large language models via discussion framework and role-play},
  author={Lu, Li-Chun and Chen, Shou-Jen and Pai, Tsung-Min and Yu, Chan-Hung and Lee, Hung-yi and Sun, Shao-Hua},
  journal={arXiv preprint arXiv:2405.06373},
  year={2024}
}

@article{sheng2024hybridflow,
  title   = {HybridFlow: A Flexible and Efficient RLHF Framework},
  author  = {Guangming Sheng and Chi Zhang and Zilingfeng Ye and Xibin Wu and Wang Zhang and Ru Zhang and Yanghua Peng and Haibin Lin and Chuan Wu},
  year    = {2024},
  journal = {arXiv preprint arXiv: 2409.19256}
}

@article{agarwal2026autodiscovery,
  title={Autodiscovery: Open-ended scientific discovery via bayesian surprise},
  author={Agarwal, Dhruv and Majumder, Bodhisattwa Prasad and Adamson, Reece and Chakravorty, Megha and Gavireddy, Satvika Reddy and Parashar, Aditya and Surana, Harshit and Dalvi Mishra, Bhavana and McCallum, Andrew and Sabharwal, Ashish and others},
  journal={Advances in Neural Information Processing Systems},
  volume={38},
  pages={25181--25219},
  year={2026}
}

@article{hirsch2006anniversary,
  title={An anniversary for cancer chemotherapy},
  author={Hirsch, Jules},
  journal={Jama},
  volume={296},
  number={12},
  pages={1518--1520},
  year={2006},
  publisher={American Medical Association}
}

@inproceedings{ligon2004penicillin,
  title={Penicillin: its discovery and early development},
  author={Ligon, B Lee},
  booktitle={Seminars in pediatric infectious diseases},
  volume={15},
  number={1},
  pages={52--57},
  year={2004},
  organization={Elsevier}
}

@article{mednick1962associative,
  title={The associative basis of the creative process.},
  author={Mednick, Sarnoff},
  journal={Psychological review},
  volume={69},
  number={3},
  pages={220},
  year={1962},
  publisher={American Psychological Association}
}

@article{ward1999creative,
  title={Creative cognition},
  author={Ward, Thomas B and Smith, Steven M and Finke, Ronald A},
  journal={Handbook of creativity},
  volume={189},
  pages={212},
  year={1999}
}

@article{runco2012standard,
  title={The standard definition of creativity},
  author={Runco, Mark A and Jaeger, Garrett J},
  journal={Creativity research journal},
  volume={24},
  number={1},
  pages={92--96},
  year={2012},
  publisher={Taylor \& Francis}
}

@article{schmidhuber2010formal,
  title={Formal theory of creativity, fun, and intrinsic motivation (1990--2010)},
  author={Schmidhuber, J{\"u}rgen},
  journal={IEEE transactions on autonomous mental development},
  volume={2},
  number={3},
  pages={230--247},
  year={2010},
  publisher={Ieee}
}

@article{mcfadzean1998creativity,
  title={The creativity continuum: Towards a classification of creative problem solving techniques},
  author={McFadzean, Elspeth},
  journal={Creativity and Innovation Management},
  volume={7},
  number={3},
  pages={131--139},
  year={1998},
  publisher={Wiley Online Library}
}

@article{boden1998creativity,
  title={Creativity and artificial intelligence},
  author={Boden, Margaret A},
  journal={Artificial intelligence},
  volume={103},
  number={1-2},
  pages={347--356},
  year={1998},
  publisher={Elsevier}
}

@article{nijstad2010dual,
  title={The dual pathway to creativity model: Creative ideation as a function of flexibility and persistence},
  author={Nijstad, Bernard A and De Dreu, Carsten KW and Rietzschel, Eric F and Baas, Matthijs},
  journal={European review of social psychology},
  volume={21},
  number={1},
  pages={34--77},
  year={2010},
  publisher={Taylor \& Francis}
}

@inproceedings{miles2017taxonomy,
  title={A taxonomy of research gaps: Identifying and defining the seven research gaps},
  author={Miles, D Anthony},
  booktitle={Doctoral student workshop: finding research gaps-research methods and strategies, Dallas, Texas},
  volume={1},
  pages={1--10},
  year={2017}
}

@article{wobbrock2012seven,
  title={Seven research contributions in HCI},
  author={Wobbrock, Jacob O},
  journal={Intelligence},
  volume={174},
  number={12-13},
  pages={910--950},
  year={2012}
}

@article{chen2026measuring,
  title={Measuring the Gap Between Human and LLM Research Ideas},
  author={Chen, Ziyu and Zhao, Yilun and Cohan, Arman},
  journal={arXiv preprint arXiv:2607.01233},
  year={2026}
}

@article{patel2026llms,
  title={Are LLMs becoming similarly creative? Evidence from three years of models},
  author={Patel, Nirav and Crossman, Josiah and Aggarwal, Eva and Wenger, Emily},
  journal={arXiv preprint arXiv:2608.19437},
  year={2026}
}
\bibliographystyle{iclr2027_conference}

\appendix

\section{Confidence Intervals for Main Results}
\label{app:main-results-ci}

\begin{table}[h]
\centering
\scriptsize
\caption{Results with 95\% confidence intervals. Citation and originality are pairwise win rates against matched reconstructed references; diversity metrics report normalized effective category coverage.}
\label{tab:main_results_ci}

\renewcommand{\arraystretch}{1.25}
\setlength{\tabcolsep}{4pt}
\resulttablerules

\begin{tabularx}{\textwidth}{
    @{}
    X
    >{\centering\arraybackslash}p{2.15cm}
    >{\centering\arraybackslash}p{2.15cm}
    >{\centering\arraybackslash}p{2.15cm}
    >{\centering\arraybackslash}p{2.15cm}
    @{}
}
\toprule

&
\multicolumn{2}{c}{\textbf{Proposal Quality}} &
\multicolumn{2}{c@{}}{\textbf{Research-Idea Diversity}} \\[-2pt]

\rowcolor{ActionBlue}
\color{white}\textbf{Method} &
\color{white}\textbf{Citation (\%)} &
\color{white}\textbf{Originality (\%)} &
\color{white}\textbf{Paradigm} &
\color{white}\textbf{Contribution} \\
\midrule

\rowcolor{ActionBlue!12}
\multicolumn{5}{@{}l@{}}{\textbf{Zero-shot}} \\

Qwen3-8B &
0.46 {\scriptsize [0.13, 1.66]} &
2.53 {\scriptsize [1.42, 4.47]} &
0.738 {\scriptsize [0.682, 0.785]} &
0.370 {\scriptsize [0.340, 0.399]} \\

\rowcolor{ResultAlt}
GPT-4.1 &
3.91 {\scriptsize [2.45, 6.17]} &
12.18 {\scriptsize [9.44, 15.59]} &
0.785 {\scriptsize [0.733, 0.826]} &
\textbf{0.439} {\scriptsize \textbf{[0.401, 0.472]}} \\

\midrule

\rowcolor{ActionBlue!12}
\multicolumn{5}{@{}l@{}}{\textbf{RL Baselines}} \\

Qwen3-8B + Temp &
11.52 {\scriptsize [8.85, 14.87]} &
19.82 {\scriptsize [16.34, 23.82]} &
0.792 {\scriptsize [0.742, 0.831]} &
0.339 {\scriptsize [0.313, 0.364]} \\

\rowcolor{ResultAlt}
Qwen3-8B + ReAct &
24.94 {\scriptsize [20.93, 29.42]} &
46.60 {\scriptsize [41.75, 51.52]} &
\underline{0.888} {\scriptsize \underline{[0.847, 0.916]}} &
0.390 {\scriptsize [0.366, 0.412]} \\

\midrule

\rowcolor{ActionBlue!12}
\multicolumn{5}{@{}l@{}}{\textbf{\textsc{AI Night-Scientist}}} \\

\textsc{Night-8B} ($R_{\text{po}}$) &
26.21 {\scriptsize [22.30, 30.53]} &
\textbf{61.61} {\scriptsize \textbf{[56.96, 66.06]}} &
0.853 {\scriptsize [0.807, 0.887]} &
0.362 {\scriptsize [0.338, 0.384]} \\

\textbf{\textsc{Night-8B} ($R_{\text{out}}$)} &
\textbf{29.89} {\scriptsize \textbf{[25.77, 34.35]}} &
56.32 {\scriptsize [51.63, 60.91]} &
\textbf{0.943} {\scriptsize \textbf{[0.907, 0.962]}} &
\underline{0.425} {\scriptsize \underline{[0.412, 0.435]}} \\

\bottomrule
\end{tabularx}

\normaltablerules
\end{table}

\par Table~\ref{tab:main_results_ci} reports 95\% confidence intervals for the primary proposal-quality and research-idea diversity results shown in Table~\ref{tab:main_results}. Citation and originality are pairwise win rates against matched reconstructed reference proposals; their intervals are Wilson score intervals over the evaluated proposal pairs, using the same treatment of ties and undecided judgments as the reported win rates. Paradigm and contribution diversity are measured as the normalized effective number of categories represented, $\exp(H)/K$, where $H$ is Shannon entropy and $K=7$; their intervals are 95\% percentile intervals from 100,000 proposal-level bootstrap resamples within each method. These intervals quantify evaluation-sample uncertainty only.

\begin{table}[h]
\centering
\scriptsize
\caption{Distribution of research domains across training proposals, grouped into high-level domain families. Fine-grained labels are derived from a three-stage automated taxonomy pipeline applied to 4,414 NSF award titles; high-level families are used only to organize the labels for presentation.}
\label{tab:domain_distribution}

\renewcommand{\arraystretch}{1.08}
\setlength{\tabcolsep}{6pt}
\nstablerules

\begin{tabularx}{\textwidth}{@{}Xrr@{}}
\toprule
\rowcolor{ActionBlue}
\color{white}\textbf{Domain} &
\color{white}\textbf{Count} &
\color{white}\textbf{\% of Labels} \\
\midrule

\rowcolor{ActionBlue!12}
\textbf{Computing \& AI} &
\textbf{6,418} &
\textbf{58.0\%} \\

Artificial Intelligence \& Machine Learning & 2558 & 23.1 \\
\rowcolor{NSTableStripe}
Networks \& Communications & 657 & 5.9 \\
Computer Systems \& Architecture & 626 & 5.7 \\
\rowcolor{NSTableStripe}
Data Science \& Analytics & 611 & 5.5 \\
Cybersecurity \& Privacy & 547 & 4.9 \\
\rowcolor{NSTableStripe}
Robotics \& Autonomous Systems & 455 & 4.1 \\
Embedded \& Hardware Systems & 310 & 2.8 \\
\rowcolor{NSTableStripe}
Cloud \& Distributed Computing & 277 & 2.5 \\
Software Engineering \& Programming Languages & 164 & 1.5 \\
\rowcolor{NSTableStripe}
Natural Language Processing \& Linguistics & 157 & 1.4 \\
Extended Reality \& Immersive Technologies & 34 & 0.3 \\
\rowcolor{NSTableStripe}
Sustainable Computing \& Infrastructure & 22 & 0.2 \\

\midrule

\rowcolor{ActionBlue!12}
\textbf{Mathematical \& Computational Foundations} &
\textbf{1,942} &
\textbf{17.5\%} \\

Computer Science Theory & 1239 & 11.2 \\
\rowcolor{NSTableStripe}
Operations Research \& Optimization & 171 & 1.5 \\
Modeling, Simulation \& Visualization & 152 & 1.4 \\
\rowcolor{NSTableStripe}
Mathematics \& Theoretical Foundations & 123 & 1.1 \\
Experimental Methods \& Evaluation & 100 & 0.9 \\
\rowcolor{NSTableStripe}
Statistics \& Data Science & 80 & 0.7 \\
Systems Science \& Engineering & 77 & 0.7 \\

\midrule

\rowcolor{ActionBlue!12}
\textbf{Human, Social \& Educational Research} &
\textbf{1,344} &
\textbf{12.1\%} \\

Human-Computer Interaction & 538 & 4.9 \\
\rowcolor{NSTableStripe}
Cognitive \& Behavioral Sciences & 241 & 2.2 \\
Education Research \& Pedagogy & 236 & 2.1 \\
\rowcolor{NSTableStripe}
Computational Social Sciences \& Digital Humanities & 173 & 1.6 \\
Career Development \& Workforce & 50 & 0.5 \\
\rowcolor{NSTableStripe}
Ethics, Equity \& Societal Impacts & 47 & 0.4 \\
Organizational Computing \& Workflow & 25 & 0.2 \\
\rowcolor{NSTableStripe}
Economics, Policy \& Law & 22 & 0.2 \\
Legal Informatics \& Law & 7 & 0.1 \\
\rowcolor{NSTableStripe}
Media, Creativity \& Inclusive Technologies & 5 & 0.0 \\

\midrule

\rowcolor{ActionBlue!12}
\textbf{Health \& Life Sciences} &
\textbf{543} &
\textbf{4.9\%} \\

Biomedical Engineering \& Health Informatics & 295 & 2.7 \\
\rowcolor{NSTableStripe}
Biological \& Biomedical Sciences & 145 & 1.3 \\
Healthcare \& Medical Sciences & 103 & 0.9 \\

\midrule

\rowcolor{ActionBlue!12}
\textbf{Engineering \& Physical Sciences} &
\textbf{532} &
\textbf{4.8\%} \\

Imaging, Instrumentation \& Sensors & 122 & 1.1 \\
\rowcolor{NSTableStripe}
Materials Science \& Nanotechnology & 103 & 0.9 \\
Quantum Science \& Engineering & 97 & 0.9 \\
\rowcolor{NSTableStripe}
Electrical \& Electronic Engineering & 63 & 0.6 \\
Energy, Power \& Automotive Systems & 46 & 0.4 \\
\rowcolor{NSTableStripe}
Safety, Risk \& Resilience Engineering & 45 & 0.4 \\
Physical Sciences & 42 & 0.4 \\
\rowcolor{NSTableStripe}
Chemistry \& Materials Science & 14 & 0.1 \\

\midrule

\rowcolor{ActionBlue!12}
\textbf{Earth, Environment \& Urban Systems} &
\textbf{193} &
\textbf{1.7\%} \\

Earth \& Geosciences & 116 & 1.0 \\
\rowcolor{NSTableStripe}
Agricultural \& Environmental Sciences & 31 & 0.3 \\
Urban Informatics \& Smart Cities & 26 & 0.2 \\
\rowcolor{NSTableStripe}
Environmental Informatics \& Sensing & 20 & 0.2 \\

\midrule

\rowcolor{ActionBlue!12}
\textbf{Research Infrastructure \& Other} &
\textbf{99} &
\textbf{0.9\%} \\

Open Science \& Research Infrastructure & 76 & 0.7 \\
\rowcolor{NSTableStripe}
Miscellaneous & 23 & 0.2 \\

\midrule
\textbf{Total} &
\textbf{11,071} &
\textbf{100.0\%} \\
\bottomrule

\end{tabularx}

\normaltablerules
\end{table}

\section{Dataset Construction \& Domain Coverage}
\label{app:dataset-details}

\subsection{Filtering and Splits}

\par We select NSF CSE awards from 2018 onwards. Non-research grants (e.g., travel and conference grants) are excluded. This yields 4,905 awards in total, split 90:10 into 4,414 training and 491 test instances. We focus on CSE awards to maximize the availability of open-access literature on arXiv, which underpins both the search action and the retrieval steps in the reward pipeline.

\subsection{Domain Distribution}

\par To characterize the topical breadth of the training set, we applied an automated three-stage taxonomy pipeline to the 4,414 NSF award titles. In the first stage, GPT-4.1 assigned one to three fine-grained research subfields to each title in batches of 50, yielding 11,071 raw subfield labels (2.5 per title on average) spanning 4,300 unique terms. In the second stage, these labels were grouped into intermediate clusters via a second LLM pass. In the third stage, the intermediate clusters were consolidated into a final canonical taxonomy of 46 research domains (Table~\ref{tab:domain_distribution}).

\par The distribution reflects the intended scope of the NSF CISE directorate: \textit{Artificial Intelligence \& Machine Learning} accounts for 23.1\% of all subfield label occurrences, followed by \textit{Computer Science Theory} (11.2\%) and \textit{Networks \& Communications} (5.9\%). The concentration in core CS and engineering is expected, especially as open-access literature on arXiv is most comprehensive for these fields, making retrieval-grounded proposal generation most reliable there. Importantly, the training set is not exclusively CS-focused: the remaining $\sim$20\% of labels span applied and interdisciplinary domains, including \textit{Biomedical Engineering \& Health Informatics} (2.7\%), \textit{Cognitive \& Behavioral Sciences} (2.2\%), \textit{Earth \& Geosciences} (1.0\%), \textit{Quantum Science \& Engineering} (0.9\%), and policy-adjacent areas such as \textit{Ethics, Equity \& Societal Impacts} and \textit{Economics, Policy \& Law}, providing a diverse training signal across 46 domains in total.


\begin{table}[h]
\centering
\footnotesize
\caption{Domain-level gains of \textsc{Night-8B} ($R_{\text{out}}$) over Qwen3-8B zero-shot, grouped into the same high-level domain families as Table~\ref{tab:domain_distribution}. Gains are differences in pairwise win rate (percentage points); uncertainty is propagated as $\sqrt{\sigma_{\text{Night}}^2 + \sigma_{\text{zs}}^2}$. Domains within each family are ordered by originality gain. ``---'' indicates insufficient samples ($<5$ pairs) for a domain-level estimate.}
\label{tab:domain_gain}

\renewcommand{\arraystretch}{1.08}
\setlength{\tabcolsep}{6pt}
\resulttablerules

\begin{tabularx}{\textwidth}{@{}Xrr@{}}
\toprule
\rowcolor{ResultHeader}
\textbf{Domain} &
\textbf{Citation Gain (pp)} &
\textbf{Originality Gain (pp)} \\
\midrule

\rowcolor{ActionBlue!12}
\textbf{Computing \& AI} & & \\

Networks \& Communications
& +40.0 $\pm$ 7.3
& +59.6 $\pm$ 7.2 \\

\rowcolor{NSTableStripe}
Robotics \& Autonomous Systems
& +34.0 $\pm$ 6.9
& +57.4 $\pm$ 7.7 \\

Artificial Intelligence \& Machine Learning
& +29.6 $\pm$ 3.4
& +57.0 $\pm$ 3.7 \\

\rowcolor{NSTableStripe}
Cybersecurity \& Privacy
& +43.2 $\pm$ 7.5
& +55.6 $\pm$ 7.7 \\

Computer Systems \& Architecture
& +30.6 $\pm$ 7.0
& +54.0 $\pm$ 7.5 \\

\rowcolor{NSTableStripe}
Data Science \& Analytics
& +28.6 $\pm$ 7.6
& +50.0 $\pm$ 9.1 \\

Cloud \& Distributed Computing
& +42.9 $\pm$ 9.4
& +50.0 $\pm$ 9.1 \\

\rowcolor{NSTableStripe}
Embedded \& Hardware Systems
& +36.1 $\pm$ 11.5
& +47.8 $\pm$ 11.3 \\

Software Engineering \& Programming Languages
& +23.1 $\pm$ 11.7
& +46.2 $\pm$ 13.8 \\

\rowcolor{NSTableStripe}
Natural Language Processing \& Linguistics
& +22.2 $\pm$ 9.8
& +36.8 $\pm$ 11.1 \\

\midrule

\rowcolor{ActionBlue!12}
\textbf{Mathematical \& Computational Foundations} & & \\

Systems Science \& Engineering
& +60.0 $\pm$ 21.9
& +60.0 $\pm$ 25.3 \\

\rowcolor{NSTableStripe}
Statistics \& Data Science
& +14.3 $\pm$ 13.2
& +57.1 $\pm$ 18.7 \\

Mathematics \& Theoretical Foundations
& +23.5 $\pm$ 10.3
& +52.9 $\pm$ 13.2 \\

\rowcolor{NSTableStripe}
Computer Science Theory
& +26.3 $\pm$ 4.7
& +49.0 $\pm$ 5.6 \\

Modeling, Simulation \& Visualization
& +31.2 $\pm$ 11.6
& +47.1 $\pm$ 13.4 \\

\rowcolor{NSTableStripe}
Experimental Methods \& Evaluation
& +36.4 $\pm$ 14.5
& +45.5 $\pm$ 15.0 \\

Operations Research \& Optimization
& +25.0 $\pm$ 12.5
& +38.5 $\pm$ 13.5 \\

\midrule

\rowcolor{ActionBlue!12}
\textbf{Human, Social \& Educational Research} & & \\

Organizational Computing \& Workflow
& +20.0 $\pm$ 17.9
& +83.3 $\pm$ 15.2 \\

\rowcolor{NSTableStripe}
Computational Social Sciences \& Digital Humanities
& +18.8 $\pm$ 9.8
& +70.6 $\pm$ 11.1 \\

Ethics, Equity \& Societal Impacts
& +0.0 $\pm$ 0.0
& +66.7 $\pm$ 15.7 \\

\rowcolor{NSTableStripe}
Education Research \& Pedagogy
& +21.4 $\pm$ 11.0
& +64.3 $\pm$ 13.9 \\

Cognitive \& Behavioral Sciences
& +47.1 $\pm$ 12.1
& +57.9 $\pm$ 11.3 \\

\rowcolor{NSTableStripe}
Human-Computer Interaction
& +29.3 $\pm$ 7.1
& +51.2 $\pm$ 8.2 \\

Career Development \& Workforce
& +0.0 $\pm$ 0.0
& +50.0 $\pm$ 20.4 \\

\midrule

\rowcolor{ActionBlue!12}
\textbf{Health \& Life Sciences} & & \\

Biomedical Engineering \& Health Informatics
& +41.2 $\pm$ 11.9
& +61.1 $\pm$ 11.5 \\

\rowcolor{NSTableStripe}
Healthcare \& Medical Sciences
& +43.8 $\pm$ 12.4
& +56.2 $\pm$ 13.5 \\

Biological \& Biomedical Sciences
& +25.0 $\pm$ 12.5
& +50.0 $\pm$ 14.4 \\

\midrule

\rowcolor{ActionBlue!12}
\textbf{Engineering \& Physical Sciences} & & \\

Quantum Science \& Engineering
& ---
& +80.0 $\pm$ 17.9 \\

\rowcolor{NSTableStripe}
Safety, Risk \& Resilience Engineering
& +35.7 $\pm$ 12.8
& +71.4 $\pm$ 12.1 \\

Materials Science \& Nanotechnology
& +7.7 $\pm$ 7.4
& +64.3 $\pm$ 13.9 \\

\rowcolor{NSTableStripe}
Imaging, Instrumentation \& Sensors
& +35.3 $\pm$ 11.6
& +61.1 $\pm$ 11.5 \\

Chemistry \& Materials Science
& ---
& +60.0 $\pm$ 21.9 \\

\rowcolor{NSTableStripe}
Energy, Power \& Automotive Systems
& +50.0 $\pm$ 17.7
& +37.5 $\pm$ 17.1 \\

\midrule

\rowcolor{ActionBlue!12}
\textbf{Earth, Environment \& Urban Systems} & & \\

Earth \& Geosciences
& +14.3 $\pm$ 13.2
& +71.4 $\pm$ 17.1 \\

\rowcolor{NSTableStripe}
Environmental Informatics \& Sensing
& +62.5 $\pm$ 17.1
& +50.0 $\pm$ 17.7 \\

\midrule

\rowcolor{ActionBlue!12}
\textbf{Research Infrastructure \& Other} & & \\

Open Science \& Research Infrastructure
& +25.0 $\pm$ 12.5
& +58.3 $\pm$ 15.8 \\

\bottomrule
\end{tabularx}

\normaltablerules
\end{table}

\section{Performance Across Research Domains}
\label{app:domain_gen}
\par We report per-domain win rates for \textsc{AI Night-Scientist-8B} and Qwen3-8B zero-shot
against the reference proposals across 36 canonical research domains
(Table~\ref{tab:domain_gain}).
The zero-shot baseline achieves near-zero impact win rates in almost every domain
(overall $0.5\%$) and near-zero originality win rates (overall $2.5\%$), confirming that
an untuned 8B model cannot produce proposals competitive with actual NSF awards on
either metric.
\textsc{AI Night-Scientist-8B} consistently overcomes this gap, delivering positive impact
gains in all 34 domains with sufficient decided pairs, ranging from $+8$ percentage points (pp)
(Materials Science \& Nanotechnology) to $+63$~pp (Environmental Informatics \& Sensing).
Originality gains are uniformly large across all 36 domains, ranging from $+37$~pp
(Natural Language Processing \& Linguistics, Energy, Power \& Automotive Systems) to
$+83$~pp (Organizational Computing \& Workflow), with particularly strong improvements
in interdisciplinary and application-oriented fields such as
Quantum Science \& Engineering ($+80.0 \pm 17.9$~pp),
Safety, Risk \& Resilience Engineering ($+71.4 \pm 12.1$~pp),
Earth \& Geosciences ($+71.4 \pm 17.1$~pp),
and Computational Social Sciences \& Digital Humanities ($+70.6 \pm 11.1$~pp).

\par Core CS domains (AI \& ML, Computer Science Theory, Robotics) show more moderate but
still substantial originality gains ($+49$--$+57$~pp), possibly reflecting that these are the most prevalent domains within the base model's pretraining distribution.
The broad consistency of gains across all 36 domains (including fields with little
CS overlap such as Biomedical Engineering, Healthcare, Materials Science, and Earth
Sciences) suggests that the improvements are not confined to a narrow research subdomain.


\section{Qualitative Analysis}
\label{app:qualitative_analysis}

\begin{table}[h]
\scriptsize
\caption{Qualitative comparison for the NSF topic \textit{``Using AI to Transform Online Video Lectures.''}}
\label{tab:qualitative_comparison_long}
\renewcommand{\arraystretch}{1.8}
\nstablerules
\rowcolors{2}{NSTableStripe}{white}
\begin{tabularx}{\textwidth}{|p{1.7cm}|X|p{4.3cm}|}
\hline
\rowcolor{ActionBlue}
\color{white}\textbf{Method} & \color{white}\textbf{Key Proposal Excerpt} & \color{white}\textbf{Assessment} \\
\hline
GPT-4.1 &
  ``Develop AI algorithms to automatically segment video lectures into coherent topics and extract key instructional elements. Design and implement \textcolor{red!70!black}{conversational agents that re-present segmented lecture content interactively}, supporting \textcolor{red!70!black}{adaptive pacing, clarifications, and multimodal delivery (text, visuals, sign language, audio descriptions)}. Empirically evaluate effectiveness and inclusivity against standard video lectures. Assess scalability and real-world deployment challenges within existing platforms.'' &
  \textbf{\textcolor{green!60!black}{Pros:}} Well-structured four-phase plan; covers accessibility and inclusivity.

  \textbf{\textcolor{red!70!black}{Cons:}} Restates existing segmentation and chatbot approaches without introducing a new mechanism; no novel technical contribution beyond integration. \\
\hline
ReAct &
  ``Develop a multimodal AI framework for real-time content adaptation leveraging visual, auditory, and physiological engagement metrics to dynamically adjust lecture content. Introduces a \textcolor{green!60!black}{`multimodal engagement score' synthesizing data across modalities}. Paired with a content adaptation engine using \textcolor{red!70!black}{reinforcement learning to optimize lecture pacing and content granularity} [no state, action, or reward space defined]. Validated via controlled A/B study with 120 participants across three engagement conditions.'' &
  \textbf{\textcolor{green!60!black}{Pros:}} Concrete novel artifact (multimodal engagement score); controlled experimental design.

  \textbf{\textcolor{red!70!black}{Cons:}} RL formulation is underspecified; experimental design is elaborate relative to the degree of technical novelty. \\
\hline
\textsc{AI Night-Scientist} (Outcome) &
  ``\textcolor{green!60!black}{`Agentify': a framework that transforms passive lectures into live, agent-mediated sessions} where AI agents interweave structured questions, personalized hints, and interactive dialogues based on real-time participant behavior. Proposes co-evolution of agent templates and lecture cognitive tiers. Randomizes 200 learners across 40 lectures (STEM, Social Science, Humanities) into four groups comparing passive, over-moderated, fixed-tier, and adaptive-tier conditions. Includes \textcolor{red!70!black}{a forced-speed-slider component with 8 granular speed settings (1x--8x)} to calibrate attention span variability.'' &
  \textbf{\textcolor{green!60!black}{Pros:}} Novel reframing of lectures as co-creative sessions rather than content delivery artifacts. Central hypothesis is testable. Ambitious but sensible multi-group experimental design.

  \textbf{\textcolor{red!70!black}{Cons:}} Some design elements appear contrived (speed-slider rationale). Cognitive tier co-evolution mechanism is underdeveloped. \\
\hline
\textsc{AI Night-Scientist} (Process + Outcome) &
  ``\textcolor{green!60!black}{Progressive Content Enactors (PCEs): virtual agents that shift from automated fidelity to intentional pedagogical maladaptation}---deliberately embedding controlled errors and ambiguity so that learners are forced to detect and resolve them, \textcolor{green!60!black}{promoting active problem-solving over passive reception}. Uses a Layered Semantic Transformative Stack (LISS) to anchor distortions to content structure. Reports \textcolor{red!70!black}{cognitive engagement score (NCIR) of $c{=}0.12$, $+3.5\sigma$ retention improvement (TAR-film, 50:1 decay), and a 2.78-$\sigma$ increase in credibility-based efficacy}---none of which are independently verifiable or grounded in standard evaluation frameworks.'' &
  \textbf{\textcolor{green!60!black}{Pros:}} Most novel direction: purposeful errors as a pedagogical tool mirrors well-established ideas (e.g., error-based learning, productive failure). PCE is a fresh, actionable concept that is clearly differentiated from prior work.

  \textbf{\textcolor{red!70!black}{Cons:}} Background section relies on unverifiable metrics and internally inconsistent citations. Evaluation framework is not grounded in standard methodology. \\
\hline
\end{tabularx}
\rowcolors{2}{}{}
\normaltablerules
\end{table}

\par We qualitatively compare proposals on the NSF topic \textit{``Using Artificial Intelligence to Transform Online Video Lectures into Effective and Inclusive Agent-Based Presentations.''} GPT-4.1 produces a well-structured but generic proposal that restates existing approaches (e.g., lecture segmentation and conversational agents) without introducing a distinguishing mechanism. ReAct adds a concrete multimodal engagement score but leaves its reinforcement learning formulation underspecified. \textsc{AI Night-Scientist-8B} (outcome-only) introduces the novel \textit{Agentify} concept---treating lectures as live, co-created sessions in which an AI agent interleaves questions and dialogues in real time, rather than optimizing pre-recorded content after the fact. The process-reward variant proposes \textit{Progressive Content Enactors} (PCEs), which deliberately introduce controlled errors and ambiguities into lectures to force students to actively resolve them (analogous to how working through a flawed proof teaches more than reading a perfect one), though its background section overreaches with unverifiable metrics. Table~\ref{tab:qualitative_comparison_long} provides excerpts and error analysis; key novel or well-specified contributions are highlighted in \textcolor{green!60!black}{green}, while vague, incremental, or questionable claims are highlighted in \textcolor{red!70!black}{red}.


\section{Reward Implementation Details}
\label{app:rewards}

\par This appendix provides the complete prompts used within the outcome-level reward pipeline described in Section~\ref{sec:outcome-reward}. The reward is computed in three stages: (1) decomposing the final proposal into atomic ideas, (2) scoring each idea for originality and relevance, and (3) scoring the execution plan for feasibility against retrieved literature.

\begin{table}[h]
\footnotesize
\caption{Creativity level descriptions for each action in the research proposal generation task. The \textit{write} and \textit{complete} actions are not creativity-enabled (fixed at level 1).}
\label{tab:action-levels}

\renewcommand{\arraystretch}{1.18}
\setlength{\tabcolsep}{6pt}
\nstablerules

\begin{tabularx}{\textwidth}{@{}p{0.8cm}X@{}}
\toprule
\rowcolor{ActionBlue}
\color{white}\textbf{$c$} &
\color{white}\textbf{Description} \\
\midrule

\rowcolor{ActionBlue!12}
\multicolumn{2}{@{}l@{}}{\textbf{\textit{search}}} \\

1 & Extremely relevant to the proposal, focusing on background information and related works based on terms extracted from the proposal title. \\

\rowcolor{NSTableStripe}
2 & Very relevant, focusing on background and related works closely related to the proposal idea. \\

3 & Somewhat relevant, focusing on background that is tangentially related to the proposal idea. \\

\rowcolor{NSTableStripe}
4 & Not very relevant; explores broader topics or concepts that may not be directly related but still provide useful context. \\

5 & Very distantly relevant; explores specific alternate perspectives, domains, or philosophical questions of general interest. \\

\midrule

\rowcolor{ActionBlue!12}
\multicolumn{2}{@{}l@{}}{\textbf{\textit{debate}}} \\

1 & Discussion with a colleague in the same specific area; structured single-turn debate focused tightly on proposal elements. \\

\rowcolor{NSTableStripe}
2 & Discussion with a colleague on related topics; structured debate focused on the proposal with minor deviation permitted. \\

3 & Debate with a peer from the same high-level field but a different topic; open-ended multi-turn format with some deviation. \\

\rowcolor{NSTableStripe}
4 & Debate with a domain-expert from a different discipline or a mixture of 2--3 experts; open-ended with broader thematic scope. \\

5 & Debate with an expert from a completely different discipline or a mixture of domain-experts; unstructured, focused on exploring diverse perspectives and high-level ideas. \\

\midrule

\rowcolor{ActionBlue!12}
\multicolumn{2}{@{}l@{}}{\textbf{\textit{spark}}} \\

1 & A minor conventional idea to challenge, very specific to the proposal's current state. \\

\rowcolor{NSTableStripe}
2 & A minor conventional idea to challenge; somewhat broad but still focused on the proposal. \\

3 & A moderate conventional idea to challenge; meaningful change questioning existing assumptions. \\

\rowcolor{NSTableStripe}
4 & A significant conventional idea to challenge; broad and exploratory, with potential field-level impact. \\

5 & A major conventional idea to challenge; very broad, with the potential to revolutionize the field. \\

\bottomrule
\end{tabularx}

\normaltablerules
\vspace{-10mm}
\end{table}

\subsection{Action Space and Creativity Levels}
\label{app:action-space}

\par Table~\ref{tab:action-levels} summarizes the creativity level descriptions ($\sC_a$) for each creativity-enabled action.

\subsection{Precedence Reward Prompt}
\label{app:novelty-prompt}

\par Each decomposed idea is scored for precedence against a paper retrieved from arXiv using the idea's keyword phrase.

\begin{tcolorbox}[promptbox]
You are a reviewer for NSF proposals who has been a professor in your field for over 20 years. Evaluate the precedence of a proposal's core idea relative to the problem statement and a related paper.\\[4pt]
Problem statement: \textcolor{red}{\{problem\}}\\[2pt]
Related paper: \textcolor{red}{\{related\_paper\}}\\[4pt]
Evaluate the following core idea for its precedence. Consider how it introduces new concepts, methodologies, or perspectives that differentiate it from the related paper or existing work in the field.\\[2pt]
Idea: \textcolor{red}{\{idea\}}\\[4pt]
Score from 1 to 5:\\[2pt]
\textbf{1}~=~Not novel. Rehash of existing work.\quad\textbf{2}~=~Slightly novel. Largely derivative.\\[1pt]
\textbf{3}~=~Moderately novel.\quad\textbf{4}~=~Very novel. Potential to advance the field.\\[1pt]
\textbf{5}~=~Highly novel. Groundbreaking; challenges existing paradigms.\\[4pt]
Output JSON: \texttt{\{"score": <int>, "explanation": "<string>"\}}
\end{tcolorbox}

\subsection{Feasibility Reward Prompt}
\label{app:feasibility-prompt}

\par Each idea's execution plan is scored for feasibility against a retrieved paper grounding the assessment in existing methodological precedent.

\begin{tcolorbox}[promptbox]
You are a reviewer for NSF proposals who has been a professor in your field for over 20 years. Evaluate the feasibility of a given core idea from a proposal. A related paper is provided to contextualize the assessment.\\[4pt]
Problem statement: \textcolor{red}{\{problem\}}\\[2pt]
Related paper: \textcolor{red}{\{related\_paper\}}\\[4pt]
Evaluate the following implementation idea for its feasibility: practicality, resources required, complexity, ethical implications, and specificity. A highly feasible proposal includes specific technical details; vague ideas should be penalized.\\[2pt]
Implementation idea: \textcolor{red}{\{idea\}}\\[4pt]
Score from 1 to 5:\\[2pt]
\textbf{1}~=~Not feasible.\quad\textbf{2}~=~Slightly feasible.\quad\textbf{3}~=~Moderately feasible.\\[1pt]
\textbf{4}~=~Very feasible.\quad\textbf{5}~=~Highly feasible and specific.\\[4pt]
Output JSON: \texttt{\{"score": <int>, "explanation": "<string>"\}}
\end{tcolorbox}

\subsection{Relevance Reward Prompt}
\label{app:relevance-prompt}

\par Each decomposed idea is scored for relevance to the input problem $p$.

\begin{tcolorbox}[promptbox]
You are a reviewer for NSF proposals who has been a professor in your field for over 20 years. Evaluate the relevance of the following idea to the given problem statement: how well it addresses the problem, aligns with its objectives, and fits within the proposed methods.\\[4pt]
Problem statement: \textcolor{red}{\{problem\}}\\[2pt]
Idea: \textcolor{red}{\{idea\}}\\[4pt]
Score from 1 to 5:\\[2pt]
\textbf{1}~=~Not relevant.\quad\textbf{2}~=~Slightly relevant.\quad\textbf{3}~=~Moderately relevant.\\[1pt]
\textbf{4}~=~Very relevant.\quad\textbf{5}~=~Highly relevant; directly aligned and comprehensive.\\[4pt]
Output JSON: \texttt{\{"score": <int>, "explanation": "<string>"\}}
\end{tcolorbox}

\subsection{Action Selection Prompt}
\label{app:action-select-prompt}

\par The following prompt is provided to the model at each trajectory step to select the next action and creativity level.

\begin{tcolorbox}[promptbox]
You are a researcher selecting the best next action for your proposal writing process. The possible actions are:\\[4pt]
\textbf{search}: Generate specific search queries to retrieve information (level: 1--5).\\[1pt]
\textbf{debate}: Set up a discussion with one or more participants on a topic (level: 1--5).\\[1pt]
\textbf{spark}: Generate a novel research idea that challenges conventional thinking (level: 1--5).\\[1pt]
\textbf{write}: Write or revise the proposal (level: 1 only).\\[1pt]
\textbf{complete}: Indicate satisfaction and end the process (level: 1 only).\\[4pt]
Level 1--2~=~focused, proposal-specific.\quad Level 4--5~=~exploratory, creativity-inducing.\\[4pt]
Target problem: \textcolor{red}{\{problem\}}\\[4pt]
Output JSON: \texttt{\{"action": "<string>", "level": <int>\}}
\end{tcolorbox}


\section{Action Prompts}
\label{app:action-prompts}

\par Each action in the \textbf{AI Night-Scientist} framework is realized through a structured LLM prompt that takes the current problem $p$, creativity level $c$, and the accumulated trajectory context as inputs, and returns a JSON output. Below we document the core prompt structure and output schema for each creativity-enabled action (\textit{search}, \textit{debate}, \textit{spark}) and the \textit{write} action.

\subsection{Search Action}
\label{app:search-action-prompt}

\par The search action generates up to five arXiv search queries. The creativity level controls query relevance: level~1 produces tightly targeted queries derived from proposal keywords; level~5 produces broad, exploratory queries spanning alternate domains, philosophical perspectives, or unrelated fields. Retrieved papers are parsed and appended to the trajectory context for use by subsequent actions.

\begin{tcolorbox}[promptbox]
You are a researcher starting a literature review for a research proposal. Generate specific search queries whose relevance depends on the creativity level (1--5):\\[4pt]
\textbf{Level 1}: Extremely relevant; queries based on terms from the proposal title.\\[1pt]
\textbf{Level 2}: Very relevant; background and related works closely tied to the proposal.\\[1pt]
\textbf{Level 3}: Somewhat relevant; tangentially related background.\\[1pt]
\textbf{Level 4}: Loosely related; broader topics that still provide useful context.\\[1pt]
\textbf{Level 5}: Very distantly relevant; alternate domains, perspectives, or philosophical questions.\\[4pt]
Problem: \textcolor{red}{\{problem\}}\quad Level: \textcolor{red}{\{level\}} (\textcolor{red}{\{level\_description\}})\\[4pt]
Output JSON: \texttt{\{"search\_queries": ["<query 1>", ..., "<query 5>"]\}}
\end{tcolorbox}

\subsection{Debate Action}
\label{app:debate-action-prompt}

\par The debate action is executed in two sequential steps: (i) generating a debate setup (participants, topics, and structure) and (ii) simulating the debate conversation. The creativity level controls participant diversity and discussion structure, ranging from a focused single-turn discussion with a same-field colleague (level~1) to an open-ended, multi-turn panel with cross-disciplinary experts (level~5). The conversation is appended verbatim to the trajectory context.

\begin{tcolorbox}[promptbox]
\textbf{Step 1 --- Debate Setup.}
You are a researcher designing a debate. The creativity level controls who you debate with, what topics you cover, and how the debate is structured:\\[4pt]
\textbf{Level 1}: Single senior colleague in the same area; structured single-turn debate on proposal specifics.\\[1pt]
\textbf{Level 2}: Peer on related topics; structured, focused on the proposal with minor deviation.\\[1pt]
\textbf{Level 3}: Peer from the same field but a different topic; open-ended multi-turn, some deviation allowed.\\[1pt]
\textbf{Level 4}: Domain-expert from a different discipline or a 2--3 expert panel; open-ended, broader scope.\\[1pt]
\textbf{Level 5}: Expert(s) from a completely different discipline; unstructured, high-level philosophical exploration.\\[4pt]
Problem: \textcolor{red}{\{problem\}}\quad Level: \textcolor{red}{\{level\}}\\[4pt]
Output JSON: \texttt{\{"debate\_participants": [\{"name": "...", "job": "...", "expertise": "..."\}],}\\
\texttt{\quad"debate\_topics": ["..."], "debate\_structure": "..."\}}
\end{tcolorbox}

\begin{tcolorbox}[promptbox]
\textbf{Step 2 --- Debate Conversation.}
You are a researcher conducting the debate as set up above. Generate a coherent, multi-turn conversation where each participant draws on their specific expertise with concrete details. Responses should not be surface level.\\[4pt]
Debate setup: \textcolor{red}{\{debate\_setup\}}\\[4pt]
Output JSON: \texttt{\{"conversation\_history": [\{"speaker\_name": "...", "speaker\_response": "..."\}]\}}
\end{tcolorbox}

\subsection{Spark Action}
\label{app:spark-action-prompt}

\par The spark action generates a ``Bit-Flip'' idea~\citep{o2025sparks}: it inverts a commonly held assumption to produce a novel research direction. The creativity level governs the scope of the challenged assumption, from a specific proposal-level constraint (level~1) to a field-reshaping paradigm shift (level~5). The Bit-Flip is appended to the trajectory context and can influence subsequent write and action-selection steps.

\begin{tcolorbox}[promptbox]
You are a researcher who has just had a sudden insight that challenges conventional thinking. Generate a Bit-Flip: identify a prevailing belief in the field (the \textbf{Bit}), invert or challenge it (the \textbf{Flip}), and distill the core conceptual leap into a single phrase (the \textbf{Spark}). Creativity level controls the scope:\\[4pt]
\textbf{Level 1}: Minor, proposal-specific assumption to challenge.\\[1pt]
\textbf{Level 2}: Minor but somewhat broader assumption; questions existing approaches in the proposal.\\[1pt]
\textbf{Level 3}: Moderate assumption; meaningful challenge to existing norms within the proposal.\\[1pt]
\textbf{Level 4}: Significant, field-level assumption; substantial departure from the status quo.\\[1pt]
\textbf{Level 5}: Major, paradigm-level assumption; potential to revolutionize the field.\\[4pt]
Problem: \textcolor{red}{\{problem\}}\quad Level: \textcolor{red}{\{level\}}\\[4pt]
Output JSON: \texttt{\{"bit": "<2--3 sentences on the status quo and its limitation>",}\\
\texttt{\quad"flip": "<novel approach or perspective, $\geq$2 sentences>", "spark": "<core phrase>"\}}
\end{tcolorbox}

\subsection{Write Action}
\label{app:write-action-prompt}

\par The write action produces or revises the complete structured proposal. It takes all accumulated trajectory context as input and outputs a structured JSON proposal. The creativity level is fixed at 1 for this action; creative expression is instead encoded in the trajectory that feeds into it.

\begin{tcolorbox}[promptbox]
You are a researcher writing or revising a research proposal based on all information gathered throughout the writing process. Create a coherent, comprehensive proposal. You may deviate from prior ideas if it improves the proposal.\\[4pt]
Problem: \textcolor{red}{\{problem\}}\quad Current proposal: \textcolor{red}{\{proposal\}}\quad Context history: \textcolor{red}{\{context\}}\\[4pt]
Output JSON:\\
\texttt{\{"proposal\_title": "...",}\\
\texttt{\quad"proposal\_summary": "concise statement of what, why, and what problems it resolves",}\\
\texttt{\quad"background\_and\_significance": "historical review of the field, what remains to be done, and how this proposal advances it",}\\
\texttt{\quad"research\_plan": [\{"phase": "...", "idea": "detailed hypothesis and contribution",}\\
\texttt{\qquad"experimental\_plan": "concrete methods, evaluation, pitfalls, fallback plans"\}]\}}
\end{tcolorbox}


\section{SciJudge Evaluation Details}
\label{app:scijudge}

\par We evaluate proposal quality using \textbf{SciJudge} (OpenMOSS-Team/SciJudge-30B) \citep{tong2026ai}, a 30B-parameter model trained to predict scientific impact via citation forecasting. For each problem instance in the test set, we compare a model-generated proposal against a matched reconstructed reference NSF proposal in a pairwise setting.

\begin{tcolorbox}[promptbox]
Today is \textcolor{red}{\{date\}}. Based on the titles and abstracts of the following two proposals A and B, determine which proposal has a higher citation count across its eventual papers.\\[4pt]
\textcolor{red}{\{timing\_assumption\}} (e.g., ``Assume Proposal B was published earlier than Proposal A.'')\\[4pt]
Show your reasoning process in \texttt{<reason>} \texttt{</reason>} tags. Return the final answer in \texttt{<answer>} \texttt{</answer>} tags. The final answer should contain only `A' or `B'.\\[4pt]
\textbf{Proposal A:}\\
Title: \textcolor{red}{\{title\_a\}}\\
Abstract: \textcolor{red}{\{abstract\_a\}}\\[4pt]
\textbf{Proposal B:}\\
Title: \textcolor{red}{\{title\_b\}}\\
Abstract: \textcolor{red}{\{abstract\_b\}}
\end{tcolorbox}

\par \textbf{Randomized A/B order.} For each pair, proposal order is randomized using a deterministic per-pair seed to eliminate position bias. The model does not know which proposal originated from the system under evaluation versus the reconstructed reference.

\par \textbf{Temporal framing.} When comparing against the synthetically reconstructed proposal (Appendix~\ref{app:reconstruction}), we instruct the judge to assume the reconstructed reference proposal was published \textit{earlier}. This grounds the comparison in a realistic temporal context: the model-generated proposal is treated as the ``newer'' proposal, which must surpass the quality of the funded reference to be preferred.

\par \textbf{Output parsing.} The model responds with chain-of-thought reasoning inside \texttt{<reason>...</reason>} tags and a final answer (A or B) inside \texttt{<answer>...</answer>} tags. We extract the answer tag; if absent or malformed, we fall back to a regex match on the first standalone `A' or `B' character.

\paragraph{Reported metric.}
We report pairwise win rate: the fraction of evaluated pairs for which \texttt{SciJudge} predicts the model-generated proposal to yield greater downstream citation impact than the matched reconstructed reference.



\section{Contribution-Type Classification Details}
\label{app:contribution-type}

\par We classify each proposal by its primary scholarly contribution using \texttt{GPT-5.4-mini} and a seven-way taxonomy adapted from \citet{wobbrock2012seven}. The contribution types are: \textit{empirical}, which produces new findings from systematically gathered or analyzed data; \textit{artifact}, which creates a novel instantiated system, tool, process, intervention, or other constructed artifact; \textit{methodological}, which introduces or refines a reusable method; \textit{theoretical}, which develops reusable concepts, models, principles, hypotheses, or frameworks; \textit{benchmark or dataset}, which contributes a reusable data or evaluation resource; \textit{survey}, which synthesizes existing work into higher-level understanding; and \textit{opinion}, which advances an evidence-grounded position intended to persuade or redirect discussion.
\begin{tcolorbox}[promptbox]
You are an expert annotator of scholarly contribution types.

Label the proposal using a domain-general taxonomy of research contributions. Do not classify by topic, domain, or technical substrate.

A contribution type is the main form of knowledge or scholarly output the proposed work will add.

\textbf{Proposal to label:}\\[3pt]
Original research topic: \textcolor{red}{\{problem\}}\\[3pt]

Proposal title:\\
\textcolor{red}{\{proposal\_title\}}\\[3pt]

Proposal summary:\\
\textcolor{red}{\{proposal\_summary\}}\\[3pt]

Background and significance:\\
\textcolor{red}{\{background\_and\_significance\}}\\[3pt]

Research plan:\\
\textcolor{red}{\{research\_plan\}}\\[4pt]

Identify the proposal's principal contribution from its proposed deliverables and research plan. Classify the central knowledge claim or deliverable rather than the topic, incidental methods, or technical vocabulary. When several contribution types appear, select the one that best captures the proposal's primary scholarly output. Use a secondary label only when a distinct second contribution is substantively central.\\[4pt]

Return JSON with exactly this shape:
\begin{verbatim}
{
  "contribution_type": {
    "primary": "<one contribution type>",
    "secondary": "<one contribution type or none>"
  },
  "confidence": <0.0-1.0>
}
\end{verbatim}
\end{tcolorbox}

\par \textbf{Distinction from research-idea paradigms.}
Contribution type captures \emph{what form of scholarly output} the proposal ultimately contributes, whereas the research-idea paradigm captures \emph{what high-level research move} is used to turn an opportunity into a proposed direction. Following the research-idea annotation setup of \citet{chen2026measuring}, our paradigm categories distinguish moves such as assumption relaxation, failure mitigation, formal derivation, empirical mapping, artifact construction, and optimization. The two axes are therefore complementary rather than redundant. For example, an artifact contribution may arise from relaxing an assumption, mitigating a failure, or optimizing resource use; conversely, a measurement-oriented research idea may ultimately contribute either empirical findings or a reusable benchmark.

\par \textbf{Primary-label selection.}
The classifier is instructed to distinguish the proposal's central contribution from methods or artifacts that merely support it. In particular, empirical contributions are defined by the new findings produced from data, while benchmark or dataset contributions are defined by the reusable resource itself. Similarly, artifact contributions center on a novel instantiated invention, whereas methodological contributions center on a reusable way of conducting research or practice. Theoretical contributions may be empirically evaluated, but the reusable concept, explanation, model, or framework must remain the principal contribution.

\par \textbf{Reported diversity.}
For our diversity analysis, we use only the primary contribution label for each proposal. We measure how broadly and evenly a model distributes its proposals across the seven contribution types using the normalized effective number of categories described in Section~\ref{sec:eval_metrics}.

\section{LLM Originality Evaluation Details}
\label{app:llm-novelty}

\begin{tcolorbox}[promptbox]
\textbf{System:} You are an experienced NSF proposal reviewer. Return valid JSON only.\\[4pt]
\textbf{User:}
You are a reviewer for NSF proposals who has been a professor in your field for over 20 years.

You are comparing two proposals that address the same award problem. Judge which proposal is more original.

Originality must be judged relative to: (1)~the other proposal; (2)~the closest retrieved paper for Proposal A; (3)~the closest retrieved paper for Proposal B. If a proposal appears more novel at first glance but is very similar to its closest retrieved paper, penalize its novelty accordingly.\\[4pt]
\textbf{Award title / target problem:} \textcolor{red}{\{problem\_title\}}\\[4pt]
\textbf{Proposal A:} \textcolor{red}{\{proposal\_a\}}\\
\textbf{Closest Retrieved Paper for Proposal A:}\\
Title: \textcolor{red}{\{paper\_a\_title\}}\quad Abstract: \textcolor{red}{\{paper\_a\_abstract\}}\\[4pt]
\textbf{Proposal B:} \textcolor{red}{\{proposal\_b\}}\\
\textbf{Closest Retrieved Paper for Proposal B:}\\
Title: \textcolor{red}{\{paper\_b\_title\}}\quad Abstract: \textcolor{red}{\{paper\_b\_abstract\}}\\[4pt]
\textbf{Criteria:} Originality of the central concepts; distance from the closest retrieved paper; whether the proposal goes beyond recombining familiar ideas; depth of novelty rather than superficial novelty; whether the proposal opens genuinely new technical directions.\\[4pt]
\textbf{Calibration examples:}
\begin{itemize}[leftmargin=1.2em,itemsep=1pt]
  \item \textit{Borderline positive:} Proposal A may look less flashy but introduces a genuinely different technical lever, representation, or problem decomposition than both Proposal B and A's closest paper.
  \item \textit{Borderline negative:} Proposal A is polished but most apparent creativity comes from extending the closest paper to a new application or evaluation setting without a clearly new core idea.
  \item \textit{Borderline negative:} Proposal A combines familiar components into a multi-phase agenda; breadth alone is not novelty if the contribution is still an expected recombination of known methods.
  \item \textit{Borderline negative:} Do not reward writing sophistication, dense terminology, or detailed milestones if the underlying idea remains close to prior work.
\end{itemize}
Return JSON only: \texttt{\{"winner": "A or B", "explanation": "brief comparative rationale"\}}
\end{tcolorbox}

\par We assess Originality using \texttt{GPT-5.1} as a judge in a pairwise setting. For each test instance, the model-generated proposal and the matched reconstructed reference proposal are each submitted to our arXiv retrieval system; the top-1 closest paper is retrieved for each using \texttt{sentence-transformers/all-MiniLM-L6-v2} embedding-based search. The retrieval query is constructed from the proposal title concatenated with the first three sentences of the proposal summary. Proposal order (A vs.\ B) is randomized per-pair using a deterministic seed to eliminate position bias.

\par The judge is instructed to assess originality relative to both the competing proposal \emph{and} the closest retrieved paper for each proposal. This prevents superficially creative proposals from scoring well if their core idea closely mirrors existing literature. The judgment emphasizes: (i) originality of the central concept, (ii) distance from the retrieved nearest neighbor, (iii) whether the proposal recombines familiar ideas or opens a genuinely new technical direction, and (iv) depth of novelty over breadth or terminological density.

\par We report the fraction of pairs in which the model-generated proposal is preferred over the matched reference (model win rate).


\section{Human Evaluation Study}
\label{app:human-eval}

\par To assess the reliability of our automated evaluation metrics, we conducted a small-scale human evaluation in which two expert annotators independently judged 33 pairwise proposal comparisons (covering all three model variants: Qwen3-8B-Base, GPT-4.1, and AI Night-Scientist-8B). Each item presented annotators with a model-generated proposal alongside a matched reconstructed reference proposal; annotators selected the preferred proposal on two axes (predicted citation impact and precedence) or indicated a tie. Both independent annotators have 5+ and 15+ years of research experience in related fields, respectively.

\subsection{Inter-Annotator Agreement}

\par Table~\ref{tab:human-eval-iaa} reports inter-annotator agreement. Agreement is high on citation impact (80.0\%, $\kappa{=}0.625$, substantial) and originality (86.7\%, $\kappa{=}0.766$, substantial), indicating that both dimensions can be reliably assessed by domain-expert annotators.

\begin{table}[h]
\centering
\small
\caption{Inter-annotator agreement across evaluation dimensions ($n{=}15$ shared items).}
\label{tab:human-eval-iaa}
\resulttablerules
\begin{tabular}{|l|c|c|c|}
\hline
\rowcolor{ResultHeader}
\textbf{Metric} & \textbf{Raw Agreement} & \textbf{Cohen's $\kappa$} & \textbf{Interpretation} \\
\hline
Impact  & 80.0\% (12/15) & 0.625 & Substantial \\
Originality & 86.7\% (13/15) & 0.766 & Substantial \\
\hline
\end{tabular}
\normaltablerules
\end{table}

\subsection{Human-LLM Agreement}

\par We also measure how well the LLM judges (SciJudge for predicted citation impact, GPT-5.1 for originality) agree with human annotators. For each annotator response, we map the human A/B choice to a \textit{model} or \textit{ground\_truth} winner using the answer key, and compare against the corresponding LLM judgment; ties and null LLM outputs are excluded. Table~\ref{tab:human-llm-agreement} reports the results pooled across both annotators.

\begin{table}[h]
\centering
\small
\caption{Human-LLM agreement on predicted citation impact and originality. Ties and unparseable LLM outputs are excluded.}
\label{tab:human-llm-agreement}
\resulttablerules
\begin{tabular}{|l|c|c|}
\hline
\rowcolor{ResultHeader}
\textbf{Metric} & \textbf{LLM Judge} & \textbf{Human--LLM Agreement} \\
\hline
Impact  & SciJudge-30B & 76.7\% (46/60) \\
Originality & GPT-5.1      & 72.4\% (43/60) \\
\hline
\end{tabular}
\normaltablerules
\end{table}

\par Agreement in the low-to-mid 70s is consistent with prior work reporting LLM-judge alignment with human raters~\citep{zheng2023judging}, and notably these human--LLM agreement rates provide additional evidence that the automated judges broadly track expert preferences at the proposal level. These results support the use of \texttt{SciJudge} and \texttt{GPT-5.1} as scalable proxies for human judgment at the proposal level.


\section{Reconstructed Reference Proposal Construction}
\label{app:reconstruction}

\par Original NSF grant proposals are not publicly available. We therefore construct \emph{reconstructed reference proposals} from evidence surrounding each funded project. These reconstructions are not intended to reproduce the original proposal text; rather, they provide standardized, evidence-grounded references for research directions that were actually funded and subsequently pursued. The proposal-generating model never receives the award abstracts, project outcomes, or associated publications used in this reconstruction.

\begin{enumerate}[leftmargin=2em, itemsep=2pt]

    \item \textbf{Collect award evidence.}
    For each award, we collect its title, abstract, project outcomes report (POR), PI names, award ID, and award period from the NSF Awards Database.

    \item \textbf{Retrieve likely associated papers by PI.}
    We use PI-first retrieval over arXiv, issuing up to 12 author-based queries using PI full names, surnames, and combinations of multiple PIs. Results are deduplicated by arXiv ID. This constrains retrieval to papers plausibly authored by the funded investigators before using topical information to determine which papers are most closely associated with the award.

    \item \textbf{Rank candidate papers using award evidence.}
    Candidate papers are first ranked using PI-author overlap, topical overlap between the award evidence (title, abstract, and POR) and the paper title and abstract, and publication timing relative to the award period. We additionally reward cases in which the award title appears directly in the paper abstract. PI overlap receives the strongest weight so that topical similarity alone cannot make an unrelated paper a strong candidate.

    \item \textbf{Verify candidates using full-text evidence.}
    For the highest-ranked candidates, we retrieve the paper text and extract the abstract, introduction, methods or approach, and experiments or results, excluding references and appendices where possible. We then rescore papers using full-text evidence, including explicit mention of the NSF award ID, NSF funding acknowledgments, PI surnames, and topical overlap with the award. We retain up to three highly ranked papers per award. If no paper passes the grounding threshold, we retain the strongest PI-matched papers with usable text so that reconstruction remains grounded in likely investigator-authored work.

    \item \textbf{Reconstruct a plausible pre-award research plan.}
    We prompt \texttt{GPT-5.1} as an expert NSF PI using the award metadata, POR, and selected papers. Crucially, the prompt treats publications as \emph{downstream evidence} of what the funded project ultimately produced and asks the model to infer a plausible pre-award research plan that could have led to those outcomes, rather than summarize or copy the papers. The model is instructed not to mention the reconstruction process and to produce the same structured format used by our generated proposals: a proposal summary, background and significance, and a multi-phase research plan specifying the central idea and experimental plan for each phase.

\end{enumerate}

\par The resulting references should therefore be interpreted as plausible reconstructions of the funded research direction, not as recovered NSF proposals. We use them only as matched references for pairwise evaluation, providing a consistent comparison point grounded in projects that were funded and subsequently pursued.

\begin{tcolorbox}[promptbox]
You are an expert NSF principal investigator reconstructing a highly realistic original NSF proposal. Use the award title, abstract, project outcomes report, PI names, and likely resulting papers to infer a plausible original proposal. Do not copy text verbatim from the papers and do not mention that the proposal is reconstructed or generated.\\[4pt]
Award title: \textcolor{red}{\{title\}}\quad PI names: \textcolor{red}{\{pi\_names\}}\quad Award period: \textcolor{red}{\{dates\}}\\[4pt]
NSF award abstract: \textcolor{red}{\{abstract\}}\\[4pt]
NSF project outcomes report: \textcolor{red}{\{por\}}\\[4pt]
Likely papers supported by this award (\textcolor{red}{\{N\}} retrieved):\\
\textcolor{red}{\{paper\_blocks\}} (title, authors, arXiv ID, evidence excerpt, processed content)\\[4pt]
Write a detailed proposal concrete enough that an evaluator can see the tasks, data, algorithms, evaluation metrics, expected results, potential pitfalls, and fallback strategies.\\[4pt]
Output JSON: \texttt{\{"proposal\_title": "...", "proposal\_summary": "2--4 dense paragraphs",}\\
\texttt{\quad"background\_and\_significance": "detailed technical review of the field",}\\
\texttt{\quad"research\_plan": [\{"phase": "...", "idea": "...", "experimental\_plan": "..."\}]\}}\\[2pt]
\textit{Requirements}: 3--5 research phases; every phase must include \texttt{phase}, \texttt{idea}, and \texttt{experimental\_plan}; content must remain consistent with the award metadata and likely papers; no markdown, commentary, or code fences.
\end{tcolorbox}


\section{Process-Level Reward Details}
\label{app:process-reward}

\par The process-level reward evaluates the quality of the agent's reasoning \textit{trajectory} rather than only its final output. Each intermediate action (i.e., all actions before the final \textit{write}) is scored along two complementary dimensions: \textbf{exploration} and \textbf{contribution}. Both scores are produced by an LLM judge, mapped from the 1--5 integer scale to $[-1, 1]$ via the centering transform $s' = (s - 3) / 2$, and averaged equally into a per-action score. The overall process reward is the mean of all per-action scores; actions that were available but unused (i.e., the trajectory is shorter than the maximum allowed) are penalized with a score of $-1$.

\subsection{Exploration Scoring}

\par The exploration dimension rewards actions that cover genuinely new ground relative to everything already explored in the trajectory. A high exploration score indicates that the action introduces a novel direction, perspective, or information source that prior actions did not cover; a low score indicates redundancy. If no prior actions have been taken, the action is considered maximally exploratory by default.

\subsection{Contribution Scoring}

\par The contribution dimension rewards actions that meaningfully shaped the final proposal's quality. Each intermediate action is scored for how much it contributed to the final proposal's precedence, feasibility, and overall quality, conditioned on prior actions (to avoid rewarding redundant actions that happen to echo an earlier high-value contribution).

\begin{tcolorbox}[promptbox]
You are a reviewer for NSF proposals who has been a professor in your field for over 20 years. Evaluate whether the current action is sufficiently different and novel compared to previous actions in the research process.\\[4pt]
Problem statement: \textcolor{red}{\{problem\}}\\[2pt]
Current action type: \textcolor{red}{\{action\_type\}}\\[2pt]
Current action output: \textcolor{red}{\{action\_output\}}\\[2pt]
Prior actions taken (ordered): \textcolor{red}{\{previous\_actions\}}\\[4pt]
Assess how meaningfully different the current action is from the previous actions. Consider:\\
\textbullet\ Does it explore a new direction or perspective not previously covered?\\
\textbullet\ Would it provide substantially different information from prior actions?\\
\textbullet\ For \textit{search}: are the queries fundamentally different from prior queries or debate topics?\\
\textbullet\ For \textit{debate}: are the topics/participants substantially different from prior debates?\\
\textbullet\ For \textit{spark}: does it challenge different assumptions than prior spark actions?\\[4pt]
Assign a low score if this action would likely generate outputs similar to prior actions; assign a high score if it explores genuinely new ground.\\[4pt]
Score from 1 to 5:\\
\textbf{1}~=~Not exploratory; highly redundant with prior actions.\\
\textbf{2}~=~Slightly exploratory; mostly overlaps with prior actions.\\
\textbf{3}~=~Moderately exploratory; some new perspectives with limited novelty.\\
\textbf{4}~=~Very exploratory; covers substantially new ground.\\
\textbf{5}~=~Highly exploratory; genuinely novel direction with minimal overlap.\\[4pt]
Output JSON: \texttt{\{"score": <int>, "explanation": "<string>"\}}
\end{tcolorbox}

\begin{tcolorbox}[promptbox]
You are a reviewer for NSF proposals who has been a professor in your field for over 20 years. Evaluate how much a single intermediate action contributed to the final proposal across three dimensions: precedence, feasibility, and overall quality.\\[4pt]
Problem statement: \textcolor{red}{\{problem\}}\\[2pt]
Final proposal: \textcolor{red}{\{final\_proposal\}}\\[2pt]
Prior actions before this one (ordered): \textcolor{red}{\{previous\_actions\}}\\[2pt]
Intermediate action type: \textcolor{red}{\{action\_type\}}\\[2pt]
Intermediate action output: \textcolor{red}{\{action\_output\}}\\[4pt]
When scoring, condition on prior actions. If this action mostly repeats prior actions with similar outputs, assign low contribution scores.\\[4pt]
Score each dimension from 1 to 5:\\
\textbf{Precedence contribution:} 1~=~No contribution or harmful; 5~=~Essential contribution to novelty.\\
\textbf{Feasibility contribution:} 1~=~No contribution or harmful; 5~=~Essential contribution to feasibility.\\
\textbf{Overall quality contribution:} 1~=~No contribution or harmful; 5~=~Essential contribution to clarity, coherence, rigor, and alignment with the problem.\\[4pt]
Output JSON: \texttt{\{"precedence\_score": <int>, "precedence\_explanation": "<string>",}\\
\texttt{\quad"feasibility\_score": <int>, "feasibility\_explanation": "<string>",}\\
\texttt{\quad"quality\_score": <int>, "quality\_explanation": "<string>"\}}
\end{tcolorbox}

\subsection{Score Aggregation}

\par For each intermediate action $i$, the combined per-action process score is $r_i = (\text{explore}_i + \text{contribute}_i) / 2$, where each component has been normalized to $[-1, 1]$. The contribution score is the mean of its three sub-components (novelty, feasibility, overall quality) after normalization. The final process reward for a trajectory of $N$ intermediate actions out of a maximum of $N_{\max}$ is:
$$R_{\text{proc}} = \frac{1}{N_{\max}} \left( \sum_{i=1}^{N} r_i \;+\; (N_{\max} - N) \cdot (-1) \right)$$
Missing actions are assigned $-1$ to penalize overly short trajectories. When combined with outcome-level rewards (the PO setting), the final reward is $R_{\text{po}} = (R_{\text{proc}} + R_{\text{out}}) / 2$.


\section{Training Hyperparameters}
\label{app:training-details}

\par We report all key hyperparameters used to train \textsc{AI Night-Scientist-8B} and \textsc{AI Night-Scientist-14B}. Training is performed with Verl \citep{sheng2024hybridflow} on $4 \times 8$ NVIDIA H100 (80~GB) GPUs (32 GPUs total). We fine-tune \texttt{Qwen3-8B-Base} and \texttt{Qwen3-14B-Base} end-to-end without LoRA adapters, as LoRA is not supported for SGLang-based rollouts in Verl.

\begin{table}[h]
\footnotesize
\caption{Training and inference hyperparameters for \textsc{AI Night-Scientist}.}
\label{tab:hparams}
\renewcommand{\arraystretch}{1.16}
\setlength{\tabcolsep}{6pt}
\nstablerules

\begin{tabularx}{\textwidth}{@{}p{0.43\textwidth}X@{}}
\toprule
\rowcolor{ActionBlue}
\color{white}\textbf{Hyperparameter} &
\color{white}\textbf{Value} \\
\midrule

\rowcolor{ActionBlue!12}
\multicolumn{2}{@{}l@{}}{\textbf{\textit{Data}}} \\
Train batch size (prompts) & 32 \\
\rowcolor{NSTableStripe}
Max prompt length & 16{,}384 tokens \\
Max response length & 25{,}000 tokens \\
\rowcolor{NSTableStripe}
Dynamic batching & Enabled \\

\midrule
\rowcolor{ActionBlue!12}
\multicolumn{2}{@{}l@{}}{\textbf{\textit{Rollout (SGLang)}}} \\
Rollouts per prompt ($n$) & 8 \\
\rowcolor{NSTableStripe}
Max agent actions per trajectory & 5 (11 max assistant turns) \\
Max tool-response length & 2{,}048 tokens \\
\rowcolor{NSTableStripe}
Sampling temperature & 0.7 \\
Max model context length & 32{,}768 tokens \\
\rowcolor{NSTableStripe}
Serendipitous swap probability & 0.5 initially; decay $\gamma=0.001$ \\

\midrule
\rowcolor{ActionBlue!12}
\multicolumn{2}{@{}l@{}}{\textbf{\textit{Actor (FSDP2)}}} \\
PPO mini-batch size (prompts) & 32 \\
\rowcolor{NSTableStripe}
Max tokens per GPU (training) & 32{,}000 \\

\midrule
\rowcolor{ActionBlue!12}
\multicolumn{2}{@{}l@{}}{\textbf{\textit{Reference Model (FSDP2)}}} \\
Parameter offload & Enabled \\
\rowcolor{NSTableStripe}
Max tokens per GPU (log-prob) & 40{,}000 \\

\midrule
\rowcolor{ActionBlue!12}
\multicolumn{2}{@{}l@{}}{\textbf{\textit{Optimization}}} \\
Algorithm & GRPO \\
\rowcolor{NSTableStripe}
KL loss coefficient & 0.001 \\
KL in reward & Disabled \\
\rowcolor{NSTableStripe}
Total training steps & 170 \\

\midrule
\rowcolor{ActionBlue!12}
\multicolumn{2}{@{}l@{}}{\textbf{\textit{Retrieval and Reward Infrastructure}}} \\
arXiv index &
\href{https://www.kaggle.com/datasets/Cornell-University/arxiv}
{Kaggle arXiv snapshot} \\
\rowcolor{NSTableStripe}
Retrieval model &
\texttt{sentence-transformers/all-MiniLM-L6-v2} \\
External reward judge & \texttt{GPT-4.1} \\
\rowcolor{NSTableStripe}
Judge temperature & 0 \\
\bottomrule
\end{tabularx}

\normaltablerules
\end{table}


\section{Action-Level Entropy and Similarity Metrics}
\label{app:action-level-metrics}

\par During training, we track two diagnostic metrics, \textbf{action-level entropy} and \textbf{action-level similarity}, that measure whether the model's actual output behavior is consistent with its selected creativity level. Neither metric is used as a reward signal; they serve purely as interpretability probes.

\par The core premise of our creativity-aligned framework is that selecting a higher creativity level should \emph{demonstrably change} how the model acts, producing outputs with higher token entropy (more lexical variety) and outputs that are more distinct from the evolving proposal (higher divergence). If a model selects level~5 but generates outputs indistinguishable from level~1, the creativity selector is decorative rather than functional. These metrics operationalize that sanity check.

\paragraph{Action-level entropy.} For each action, we collect the per-token log-probabilities from the model's generation and compute the mean Shannon entropy across top-$k$ vocabulary candidates. This is averaged over all tokens in the action output to yield a single per-action entropy score. We then compute the Pearson correlation between the action's selected creativity level (equivalently, its inverse noise level $1 - \text{noise}/\text{max\_level}$) and this entropy value across all actions in a training batch. A positive correlation indicates that higher creativity levels lead to higher-entropy, more diverse token distributions.

\paragraph{Action-level similarity.} For each action, we compute the cosine similarity between the action output's sentence embedding and the embedding of the most recently written proposal draft (or the problem statement if no draft exists yet). We again compute the Pearson correlation between the creativity level and this similarity score across the batch. A negative correlation (similarity decreases as creativity level increases) is desirable: higher creativity should yield actions that cover genuinely new territory rather than paraphrasing what has already been written.


\section{Limitations}
\label{app:limitations}

\paragraph{Computational cost.} Training creativity-aligned agents requires substantial GPU compute. We mitigate this by working with an efficient 8B-parameter architecture, but scaling to larger models would amplify the footprint. Additionally, our reward pipeline relies on LLM judges (GPT-4.1 for feasibility/relevance during training, SciJudge-30B and GPT-5.1 for evaluation). Unlike similarity-based reward signals that compare generated text against existing corpora, we deliberately use LLM judges to assess genuinely \textit{new} ideas---minimizing data leakage and preventing the model from learning to simply paraphrase known work from our dataset, which spans proposals as far back as 2018. This design choice comes at higher inference cost but is central to the validity of the precedence and originality signal.

\paragraph{Training stability.} Given the inherent complexity of our task, particularly the implicit objective of increasing the entropy of a model's outputs, we have observed that training can become unstable, especially over longer training runs. High-entropy generation is at odds with the stability assumptions underlying standard RL algorithms designed for low-variance, verifiable reward settings. We hope future work will explore more robust RL algorithms that can handle the entropy increases required for creative reasoning. We also note that our infrastructure relies on \textsc{Verl} as the training backend and \textsc{SGLang} as the inference backend; their current joint support for structured, multi-step agentic frameworks is limited, and tighter integration would enable more scalable and reliable training pipelines.


\end{document}